\pdfoutput=1
\documentclass[letterpaper]{article} 
\usepackage{aaai24}  
\usepackage{times}  
\usepackage{helvet}  
\usepackage{courier}  
\usepackage[hyphens]{url}  
\usepackage{graphicx} 
\usepackage{booktabs}
\usepackage{natbib}  
\usepackage{caption} 
\usepackage{algorithm}
\usepackage{algorithmic}

\usepackage{newfloat}
\usepackage{listings}
\DeclareCaptionStyle{ruled}{labelfont=normalfont,labelsep=colon,strut=off} 
\floatstyle{ruled}
\newfloat{listing}{tb}{lst}{}
\floatname{listing}{Listing}
\title{Image Content Generation with Causal Reasoning}
\author{
	Xiaochuan Li\textsuperscript{\rm 1,4},
	Baoyu Fan\textsuperscript{\rm 2,1}\thanks{Corresponding author.},
	Runze Zhang\textsuperscript{\rm 1},
	Liang Jin\textsuperscript{\rm 1},
	Di Wang\textsuperscript{\rm 1}\\
	Zhenhua Guo\textsuperscript{\rm 1},
	Yaqian Zhao\textsuperscript{\rm 1},
	Rengang Li\textsuperscript{\rm 3,1}\\
}
\affiliations{
    \textsuperscript{\rm 1}Inspur Electronic Information Industry Co.,Ltd.
	\textsuperscript{\rm 2}Nankai University\\
	\textsuperscript{\rm 3}Tsinghua University
	\textsuperscript{\rm 4}Shandong Massive Information Technology Research Institute
}

\usepackage{bibentry}

\begin{document}

\maketitle

\begin{abstract}
The emergence of ChatGPT has once again sparked research in generative artificial intelligence (GAI).
While people have been amazed by the generated results, they have also noticed the reasoning potential reflected in the generated textual content. 
However, this current ability for causal reasoning is primarily limited to the domain of language generation, such as in models like GPT-3.
In visual modality, there is currently no equivalent research.
Considering causal reasoning in visual content generation is significant.
This is because visual information contains infinite granularity.
Particularly, images can provide more intuitive and specific demonstrations for certain reasoning tasks, especially when compared to coarse-grained text.
Hence, we propose a new image generation task called visual question answering with image (VQAI) and establish a dataset of the same name based on the classic \textit{Tom and Jerry} animated series. 
Additionally, we develop a new paradigm for image generation to tackle the challenges of this task. 
Finally, we perform extensive experiments and analyses, including visualizations of the generated content and discussions on the potentials and limitations.
The code and data are publicly available under the license of CC BY-NC-SA 4.0 for academic and non-commercial usage.
The code and dataset are publicly available at: \url{https://github.com/IEIT-AGI/MIX-Shannon/blob/main/projects/VQAI/lgd_vqai.md}.
\end{abstract}

\section{Introduction}\label{Introduction}

AI-generated content (AIGC), also known as generative AI (GAI), recently gained a surge of development \cite{zhang2023one,zhang2023complete,cao2023comprehensive,balaji2022ediffi}, covering several areas such as image \cite{ramesh2021zero,ramesh2022hierarchical,saharia2022photorealistic,yu2022scaling,rombach2022high}, text \cite{raffel2020exploring,radford2018improving,radford2019language,brown2020language,2303.08774,vinyals2015show}, 3D \cite{fu2022shapecrafter,jahan2021semantics,liu2022towards,mildenhall2021nerf}, and speech \cite{qian2014training,ze2013statistical,zen2015unidirectional}.
Since ChatGPT emerged, people have been amazed by its performance while recognizing the reasoning potential in the generated text content \cite{bang2023multitask}.
In particular, some recent studies have started to delve into the text's reasoning ability \cite{kojima2022large,wang2023plan,wei2022chain,shum2023automatic}, including causal reasoning \cite{fengshenbang,Fengshenbanglm}, in GAI.

\begin{figure}[ht]
	\centering
	\includegraphics[width=0.8\linewidth]{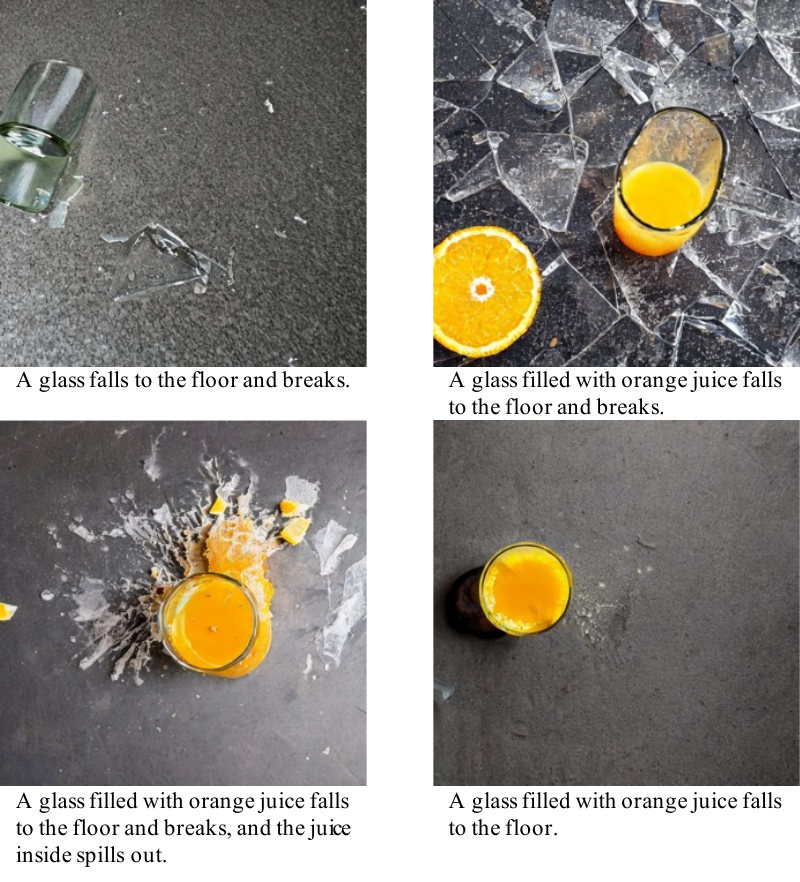}
	\caption{Generated results of Stable Diffusion v2.1. The generated results strictly follow the guidance of the text, ignoring other content caused by the implied conditions.}\label{fig1_reasoning_lack}
\end{figure}

However, majority of these works have primarily focused on text content generation, with only a limited number of studies exploring other modalities like images.
Although some studies have tried to use images as input and achieved good output results, except for some scalable vector graphics (SVG) representation for sketches or doodles \cite{2303.08774}, this field has been scarcely studied.
AIGC is currently evolving towards making the generated content more realistic. More specifically, these generated contents cover as many requirements as possible in the guidance and present more realistic details that amaze the human eye or ear.
However, these generative models are difficult to follow when underlying cause-and-effect logic is implicit in prompts like an implied condition or relationship between objects.

\begin{figure}[ht]
	\centering
	\includegraphics[width=0.9\linewidth]{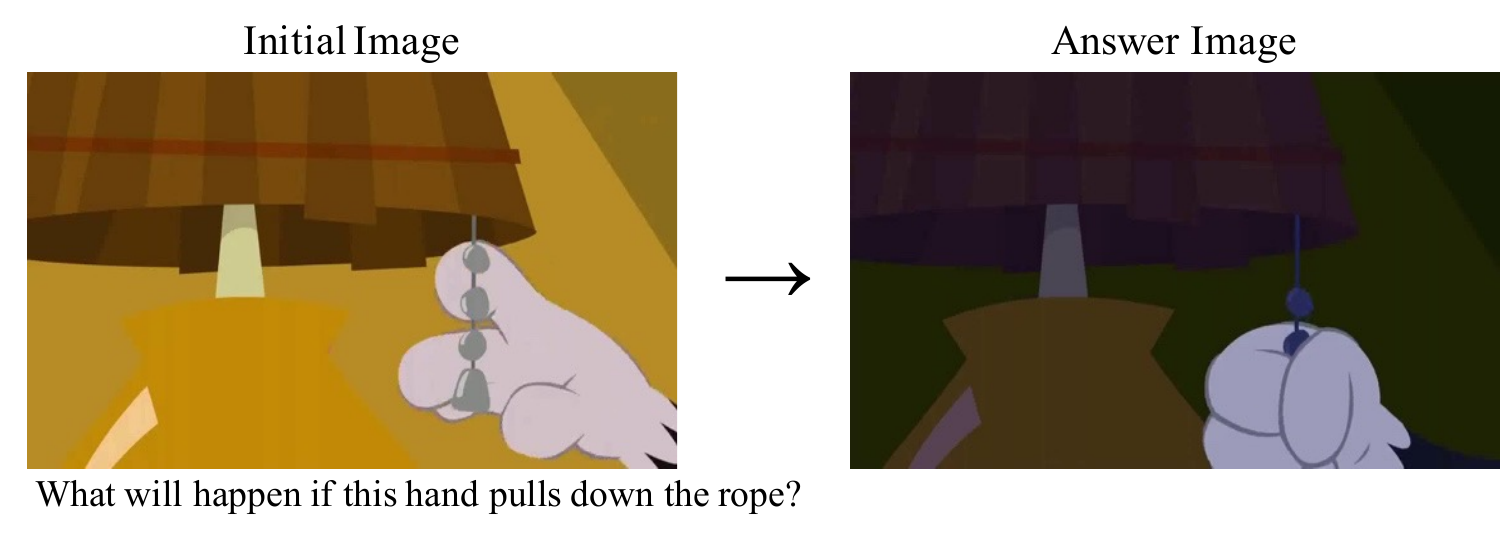}
	\caption{Task definition of VQAI.}\label{fig2_task}
\end{figure}

Regarding the image AIGC solely, popular models do not exhibit satisfying reasoning abilities.
As shown in Figure \ref{fig1_reasoning_lack}, the model is competent when we ask for a broken glass.
Once the ``filled with orange juice'' condition is added, a hidden fact is that ``the juice will be spilled out'' since ``the glass breaks''. Unfortunately, the generation fails.
However, it can be generated smoothly if we write this fact obviously into the guidance, as shown in the lower left corner.
Finally, we give only the events without any prompts of possible outcomes. The generated image does not even include the ``broken'' statement.

In the field of visual content generation, it is valuable to consider the ability of causal reasoning during image generation.
Specifically, since images contain information at infinite granularity, images can give a more intuitive and specific demonstration for some inference tasks, especially in comparison with the coarse-grained text.
In this study, we consider image content generation with causal reasoning. 
Thus, we propose a new task as shown in Figure \ref{fig2_task}. 

In particular, this task is somewhat similar to the image-editing task in terms of the form of input and output. 
More specifically, the difference is that we do not provide an exact description of the differences used to edit, but rather a text question containing a condition. 
Furthermore, models need to generate appropriate image content based on the implicit cues in the text and initial image.
From this perspective, our work can also be seen as an extension of the classical multimodal task visual question answering (VQA) \cite{antol2015vqa} in terms of output modality. 
Thus we also refer to this task as visual question answering with image (VQAI).

Accordingly, we make a new dataset for this task based on the classic \textit{Tom and Jerry} cartoon series for two main reasons.
First, \textit{Tom and Jerry} has a more straightforward worldview than the natural scenarios, meaning that the causal relationship between characters is more straightforward and clear, with very few indirect causal events like \textit{emotional hiding} and \textit{complex strategies}.
Second, it gives more prominence to the visual aspect of behaviors. Particularly, it weakens speech as much as possible and describes relationships through visual states like movements and expressions.
Besides, due to the animation, the variations of objects and backgrounds are relatively controllable, facilitating our first attempt at this task.

Particularly, we develop a new method for this task. 
An obvious idea is concatenating a multimodal comprehension module and a visual generator. 
While the former is used to generate sensible text, the latter performs image editing based on the former's output.
The text acts as a bridge in this pipeline.
However, this exposes a considerable risk - there is so much less information in text than in an image.
Consequently, it would take an enormous amount of text to replace the content in the image.
This is most likely beyond the comprehension capability of the editing model and even beyond the token length limit.
Moreover, making the language model generate long enough text is complex.
Therefore, we propose a hidden space-guided causal image generation method and conduct extensive experiments on the proposed dataset to demonstrate the scheme's effectiveness.

We summarize our main contributions as follows:

\begin{itemize}
\item{We rethink image AIGC with causal reasoning and propose a new task called VQAI.}
\item{Additionally, a new dataset is proposed to support the study of causal image generation.}
\item{Furthermore, we analyze the challenges of this task and propose a new approach to solve it. Extensive experiments demonstrate the effectiveness of our method.}
\end{itemize}

\section{Related Work}\label{Related Work}

\noindent{\textbf{\textit{Image AIGC:}}}
Image generation is an important research area of visual AIGC that drawn huge interest among researchers \cite{zhang2023one,zhang2023complete,cao2023comprehensive,balaji2022ediffi}.
In recent years, underlying generative models have been continuously proposed to promote development in this field.
Variational auto-encoder (VAE) \cite{1312.6114} is an auto-encoder that learns data distribution from latent space, and it can change the generated image by verifying the input encoding.
The generative adversarial network (GAN) \cite{goodfellow2020generative,dhariwal2021diffusion} trains a discriminator and a generator, respectively, based on the deep network to achieve automatic image generation, driving a wave of research trends.
PixelRNN \cite{van2016pixel} generates reasonable visual patches based on prior pixel sequence.
More recently, the diffusion model \cite{ho2020denoising} learns information degradation due to noise and generates images systematically using the learned patterns.

Meanwhile, in the past two years, text-guided image generation (text-to-image) has become popular with the rise of multimodal research.
DALL-E \cite{ramesh2021zero} uses a pre-trained discrete variational auto-encoder (dVAE) to extract tokens for the image and an auto-regressive transformer to generate the image.
DALL-E2 \cite{ramesh2022hierarchical} achieves good results using CLIP \cite{radford2021learning} as a text encoder and direct access to a diffusion model.
Stable/Latent Diffusion \cite{rombach2022high} replaces the image with encoded features as a supervised signal and restores them to images via a visual decoder.
Imagen \cite{saharia2022photorealistic} cascades super-resolution models after the image generator to generate high-quality, large-scale images.
Besides, Parti employs the auto-regressive model and further improves image quality by expanding the model size to 20 billion.
Furthermore, eDiffi \cite{balaji2022ediffi} generates better images by integrating expert denoisers in the diffusion model, using both CLIP \cite{radford2021learning} and T5 \cite{raffel2020exploring} as text encoders.

Meanwhile, some other tasks for image content generation are also derived.
Image editing aims to edit an image according to a given text or another description form.
Imagic \cite{kawar2022imagic} performs various text-based semantic editing on a single image, including highly complex non-rigid changes like pose changes and editing multiple objects.
InstructPix2Pix \cite{brooks2022instructpix2pix} introduces small structural changes to the diffusion model and fine-tunes it to gain editing capabilities.
More so, the story continuation task generates subsequent images based on the initial image and plot synopsis.
StoryDALL-E \cite{maharana2022storydall} uses a pre-trained text-to-image transformer to make the plot of the generated content more coherent.
AR-LDM \cite{pan2022synthesizing} trains a hidden diffusion model to improve the generated images' story continuity and content consistency.

These works promote research in visual AIGC, where image content becomes increasingly controllable and realistic.
However, these tasks require images to be generated strictly following text guidance, ignoring the ability to reason during generation.
Therefore, this work aims to develop such a topic and investigate causal image content generation.

\noindent{\textbf{\textit{Reasoning in GAI:}}}
Recently, with the birth of ChatGPT, the research for large language models (LLMs) continued to grow in popularity. Particularly, many researchers have started to explore the reasoning abilities embedded in LLMs.
It is found that adding encouragement to prompts drives reasoning in the generated text \cite{wei2022chain}.
Zero-Shot-CoT \cite{kojima2022large} concatenates ``Let's think step by step'' after a question to get a more detailed reasoning step and achieves better performance on question answering (QA) tasks.
Manual-CoT \cite{wei2022chain} manually designs a few question-answer samples to guide the language model to continue the chain of thoughts.
Auto-CoT \cite{zhang2022automatic} automatically constructs examples via Zero-Shot-CoT and demonstrates the generation of intermediate reasoning steps and the final answer.
Automatic-CoT \cite{shum2023automatic} constructs a candidate pool of rationale chains based on a small-labeled dataset and selects the best combination for CoT prompting by employing a variance-reduced policy gradient strategy.
Besides, some researchers have started to analyze the causal/counterfactual reasoning ability embodied in LLMs, such as Randeng-Deduction \cite{fengshenbang,Fengshenbanglm}.
These works illustrate that LLM-based GAI exhibits interesting reasoning capabilities, at least in text generation.

In the multimodal domain, MM-CoT \cite{zhang2023multimodal} transfers the CoT to image-text samples and enables detailed rationale and answer generation by fine-tuning an LM.
GPT4 \cite{2303.08774} demonstrates the results of causal reasoning with images and can even generate scribbles of simple images in SVG representation.
However, although these works considered images in GAI, they did not include images as outputs.
This study refers to related works in the field of LLM and multimodality to further investigate the causal capabilities in image generation.

\section{Visual Question Answering with Image}\label{Datasets}

This section presents the dataset with the same name for the proposed task VQAI.
We make the code to access the dataset publicly available under a CC BY-SA 4.0 license. 

Furthermore, we disclose more details and analysis about the dataset in \textbf{\textit{Supplementary Material}}.

\subsection{Task Definition}

\textit{Syllogism} \cite{smiley1973syllogism} is a basic unit of causal reasoning and is divided into three parts: \textit{major premise}, \textit{minor premise} and \textit{conclusion}. The \textit{major premise} is the statement of a general or universal nature. The \textit{minor premise} is the statement about a particular case. The \textit{conclusion} is a corollary to accepting the premises.
To study causal reasoning in the visual task, we use this \textit{syllogism} form to formulate the task.

\begin{figure}[ht]
	\centering
	\includegraphics[width=1.0\linewidth]{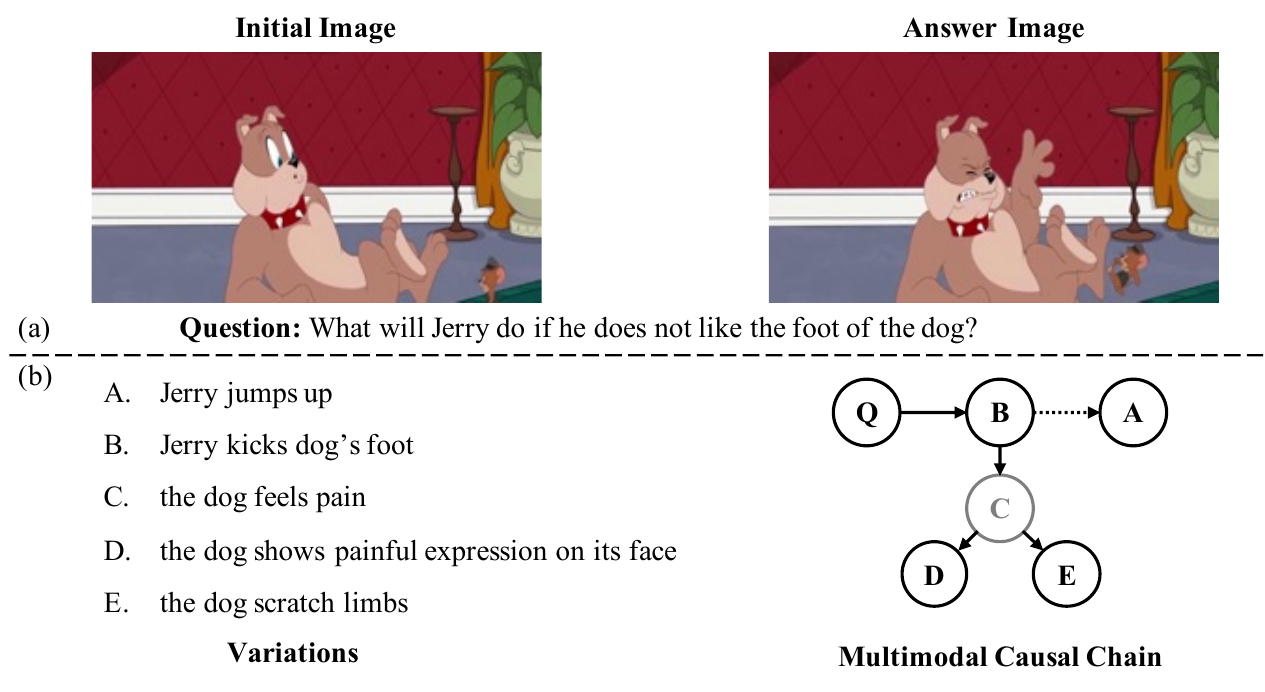}
	\caption{A sample from the VQAI dataset.}\label{fig3_dataset_sample}
\end{figure}

In particular, as shown in Figure \ref{fig3_dataset_sample} (a), we construct a sample comprising three parts: i) an initial image as the major premise, which is used to describe the relationship between the current scene and the objects; ii) an interrogative/question as the minor premise, containing a causal condition in the current scenario; and iii) an answer image as the conclusion describing a reasonable result considering both constraints.
This formulation is like a sample of VQA; thus, our work can also be seen as an extension of VQA considering causal reasoning on the output modality.

\subsection{Data Collection}

\begin{figure*}[ht]
	\centering
	\includegraphics[width=0.87\linewidth]{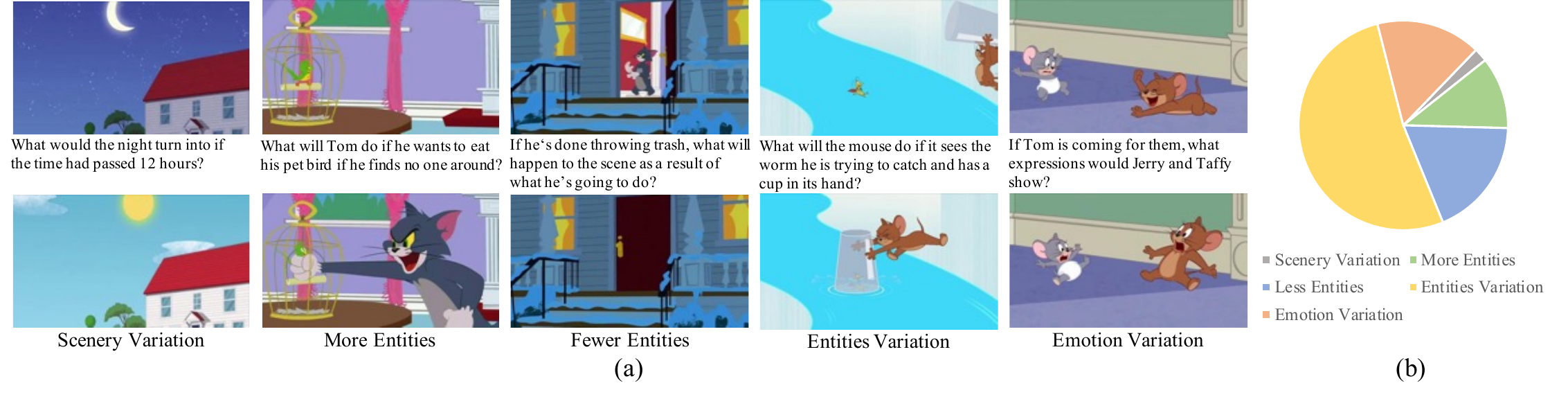}
	\caption{Demonstration and proportion of the five categories of samples in the dataset.}\label{fig4_5_five_data}
\end{figure*}

For the prompt VQAI task, we produce a new dataset.
All pictures used in the dataset are sampled from the \textit{Tom and Jerry} cartoon series.
We adopt this cartoon for several main reasons.
First, compared to the complex scenes of nature, the worldview of cartoons is often simplified. 
Specifically, the relationships between different entities are greatly simplified in the world of \textit{Tom and Jerry}, and there are very few overly complex events such as ``emotional hiding'' and ``complex strategies''. 
Instead, the exaggerated drawing style usually highlights or emphasizes the character's reaction to a particular condition. 
This means that the cause-and-effect relationship between characters is more straightforward and clear, facilitating our analysis and study of the task.
Second, \textit{Tom and Jerry} attaches great importance to the presentation of visuals. Compared with some other animated films, this cartoon ignores the necessity of language as much as possible. 
Particularly, it tends to convey the moods and reactions of its characters through expressions, movements, or states. This is beneficial to our exploration of image causal generation.
Besides, as an animated film, the changes in backgrounds, characters, and objects are relatively controllable. 
This facilitates our first attempt at this novel task. 
Meanwhile, it reduces the difficulty of data collection.

Specifically, we download 755 episodes of \textit{Tom and Jerry} from public sources, hand-crop pairs of images where causal relationships exist, and label the interrogative sentences that contain the conditions.

\subsection{Annotations}
\noindent{\textbf{\textit{Causal Questions:}}} 
The annotators are asked to annotate a causal question of the text on each pair of causal images.
Particularly, the question needs to give conditions under which an initial image evolves rationally, and the answer image must be a reasonable result of the evolution of the initial image under this condition.
An example is shown in Figure \ref{fig3_dataset_sample} (a).
In detail, each image pair is first checked by the annotator for the existence of valid causal relationships, and then these relationships are summarized in a single question, such as ``What happens if [condition]?'', ``What will it do if [condition]?'' and so on.

\noindent{\textbf{\textit{Causal Chain:}}} 
The annotation of causal questions completes the \textit{syllogism}.
However, this task is still challenging, especially as some events develop, requiring multiple reasoning steps.
It is indeed difficult for image generation models to answer complex causal questions (which is analyzed further in Section \ref{Experiments}).
Moreover, models cannot be assessed whether they truly learn causal reasoning capabilities or just fit the statistical bias on the dataset.
Therefore, we annotate the steps of reasoning, called causal chains, for a part of the samples. 
It shows how the first image develops systematically into the second under the conditions given in the causal question.
In the causal chain, each edge represents one inference step, and each node represents a variation of the event development, as shown in Figure \ref{fig3_dataset_sample} (b).

In allow the inference process to be better structured, we classify edges and nodes separately. Specifically, edges are classified into two types: i) to express causal reasoning, for example, ``Jerry kicks dog's foot'' \textbf{causes} ``the dog feels pain'', which is conventional forward reasoning; ii) to express the condition or need, for example, ``Jerry kicks dog's foot'' \textbf{needs} ``Jerry jumps up'', which looks more like a reverse thinking process out the necessary conditions.
In addition, nodes are also divided into two types, which are i) \textbf{visible} and ii) \textbf{invisible} in the image. For example, ``feeling pain'' is actually a mental activity that is not visible. However, it leads to a ``painful expression'', which is visible.

\subsection{Quality control}

We follow strict control rules to ensure the quality of the dataset, reflected in two main aspects – \textit{annotation guidance} and \textit{annotation checking}.
In particular, this quality control procedure applies to the causal questions since this part of the data is labeled by an external worker due to its large volume. Conversely, the causal chains are all labeled by three experienced researchers.

\noindent{\textbf{\textit{Annotation Guidance:}}}
We provide annotators with five strict templates to select image pairs from the video and write causal questions. 
In short, given an image pair, it is valid when and only if it satisfies one of the following rules:
\begin{itemize}
\item{\textbf{Scenery Variation}: the scene or environment is modified, such as changes in weather, brightness, and season.}
\item{\textbf{More Entities}: the scene has not been modified, but one or more entities have been added.}
\item{\textbf{Fewer Entities}: the scene has not been modified, but one or more entities have been reduced.}
\item{\textbf{Entities Variation}: the modifications to the scenario are minor with no additions or subtractions of entities.}
\item{\textbf{Emotion Variation}: one or more characters' emotions change, accompanied by expressions or movements.}
\end{itemize}


Figure \ref{fig4_5_five_data} (a) shows specific examples of these five sample categories, whose proportions are represented as shown in Figure \ref{fig4_5_five_data} (b).

\noindent{\textbf{\textit{Annotation Checking:}}}
The researchers review each causal question label upon submission consistent with the above criteria, and samples with unreasonable causal questions are rejected. 
Ultimately, VQAI contains 17,524 samples, of which 3,809 sets include causal chain annotations.

\section{Latent Guided Diffusion Model via Frozen Large Language Model}\label{Method}

\subsection{Latent Guided Image Generation}\label{lst}

\begin{figure}[ht]
	\centering
	\includegraphics[width=1.0\linewidth]{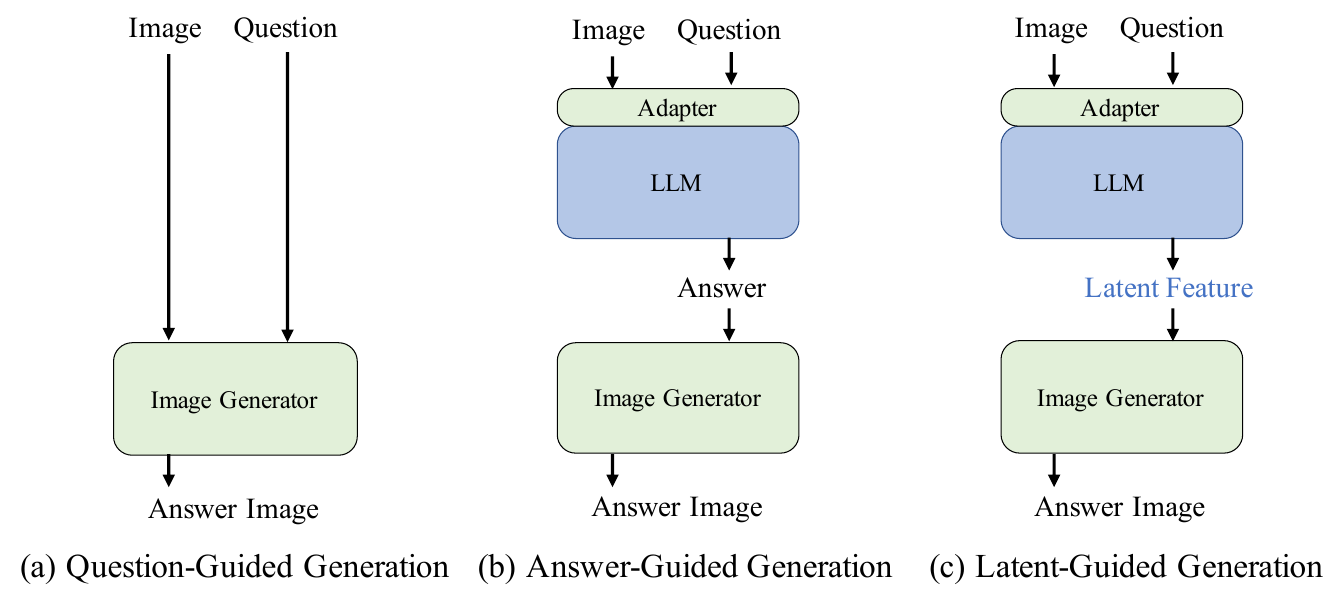}
	\caption{Three paradigms for causal image generation.}\label{fig6_three_paradigm}
\end{figure}

We think about causal reasoning and image content generation.
Specifically, one of the most straightforward solutions is to use an off-the-shelf image editing model, such as InstructPix2Pix, if only the matching of the model to the data structure is considered.
We refer to this approach as question-guided generation, as shown in Figure \ref{fig6_three_paradigm} (a).
However, since the input text does not contain explicit information for modification, it is risky to use only an image editor.
It may not be able to causal reason.
Therefore, we consider cascading a reasoning module before the generator.

The reasoning ability of large language models (LLMs) is widely recognized.
In the multimodal domain, some approaches have inserted adapters in LLM and verified that LLM maintains reasoning ability over multimodality on tasks such as VQA and image captioning \cite{li2023blip}.
Therefore, it is worth borrowing the inference capability of LLM in the causal image generation process.
In this work, we consider two different paradigms, as shown in Figure \ref{fig6_three_paradigm} (b) and (c), and refer to them as answer-guided generation and latent-guided generation, respectively.
One is to use LLM to reason about the textual output for multimodal inputs and use that answer to guide the editing model.
However, since images are far more information-rich than text, this may introduce new risks.
It requires considerable textual description to replace the equivalent amount of information in an image.
This may exceed the editing model's comprehension capability and even the token length limit.
In this study, we propose a new generative paradigm that uses the encoding features of LLM to guide the generation model.

\begin{figure*}[ht]
	\centering
	\includegraphics[width=0.87\linewidth]{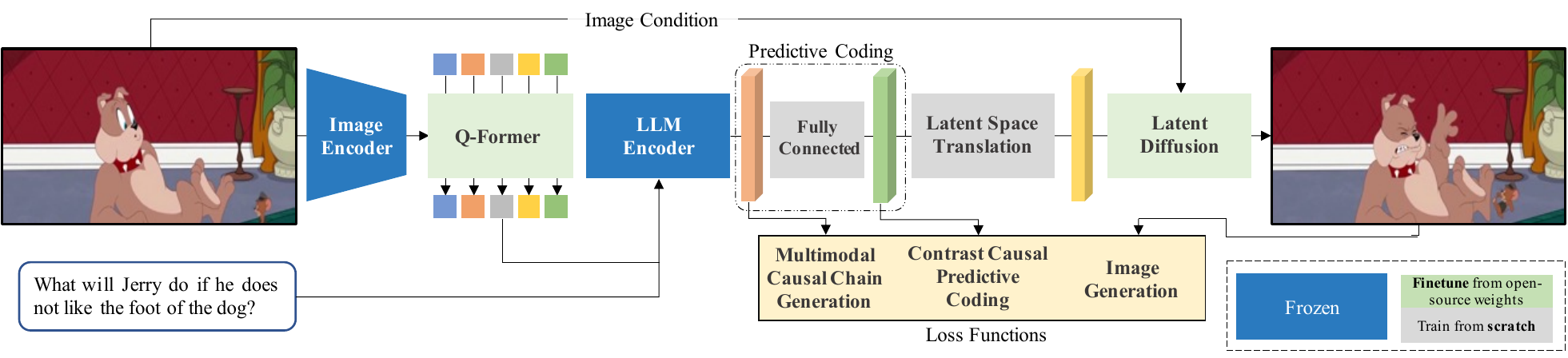}
	\caption{Architecture of Latent Guided Diffusion Model via Frozen Large Language Model.}\label{fig7_arch}
\end{figure*}

We propose a new method based on LLM with a diffusion model called latent guided diffusion (LGD), and the method structure is shown in Figure \ref{fig7_arch}.
For the use of LLM, we introduce the Q-Former in the BLIP2 \cite{li2023blip}.
Q-Former initializes a set of fixed-length query tokens and translates the image information into features that the LLM encoder can read by making cross attention to the image features.
These features are concatenated together with the embeddings of language prompts (interrogatives or other forms of instances) to implement different downstream tasks.
As shown in the figure, we use the same form of extracting multimodal features for images and causal interrogatives and use them to guide image decoding.
On the image decoding side, we refer to the related work of Stable Diffusion \cite{rombach2022high} and InstructPix2Pix \cite{brooks2022instructpix2pix} to fuse the features used for guidance into different stages of UNet through the attention mechanism.

However, this brings new challenges.
First, the diffusion model does not recognize the output features of LLM.
Since it is costly to construct a large causal dataset, we intend to refrain from training either of them from sketch.
Therefore, a space translation for latent features is necessary.
It adapts the feature's dimensionality and semantics of the feature.
In the structure, we add a fully connected layer in front of the latent diffusion model to translate the distribution of its input.

Moreover, for causal content generation, the output image is essentially a prediction of the subsequent of the initial image under a particular condition.
In other words, the features used for guidance need to express the information after being predicted.
Unfortunately, we cannot be certain that the output of the LLM encoder can explicitly contain this.
Usually, these features are inputted into the LLM decoder, and the prediction is made gradually by autoregression.
Therefore, we add a new module for predictive encoding to predict the subsequent steps triggered by causal interrogatives.
We realize this through the inspiration of predictive coding (PC) \cite{aitchison2017or,huang2011predictive,oord2018representation}.
In sequence-prediction tasks such as speech, ordered image, and video, PC predicts the prediction space after a given moment based on a given sequence by adding a new series of fully connected layer combinations.
This is very similar to the task of causal image prediction, so we transform the output of the LLM encoder into predicted information by setting up fully connected layers and encoding the prediction.

\subsection{Contrast Causal Predictive Coding}\label{ccpc}

Contrast predictive coding (CPC) \cite{oord2018representation} takes the speech or other ordered fragments as input to the encoder to extract ordered features.
The features are input to multiple isomorphic predictive coding networks with different weights to obtain predictive features for multiple moments after that fragment.
This approach induces potential space to capture valuable information to predict future samples.
Ultimately, CPC optimizes the model parameters by constraining the predictive features to the ground truth of the corresponding batch's corresponding moments.

We refer to this form because causal image generation can also be seen as a prediction task. 
Moreover, the labeled form of the causal chain appears to satisfy the conditions to construct the loss function.
However, two risks arise.
First, the time-series samples to which this predictive encoding applies are uniform.
In other words, the distance between any adjacent frames in the sequence is the same as in a speech or video sequence with a fixed time interval.
However, such uniformity does not exist in causal inference.
In particular, it is not guaranteed that all neighboring nodes of a causal chain express an equal number of inference steps between them.
As in Figure \ref{fig3_dataset_sample} (b), we consider that ``mouse kicks dog's foot'' (node \textit{B}) causes ``dog feels pain'' (node \textit{C}), followed by ``dog shows a painful expression''(node\textit{D}).
However, it seems reasonable to derive \textit{D} directly from \textit{B}. People express the inference differently, so the causal chain is not uniform.
Moreover, the annotation of causal chains has a significant long-tail effect.
This may lead to insufficient training of the latter fully-connected layers in traditional PC.

Therefore, we propose contrast causal predictive coding (CCPC).
First, we replace several FCs in CPC with one, which is only used to encode whether there is a causal relationship between the two.
Specifically, while calculating the loss, we take positive samples from the causal chain of the current sample and select several nodes from other samples as negatives, and optimize the model parameters using contrast learning.

\begin{figure}[ht]
	\centering
	\includegraphics[width=1.0\linewidth]{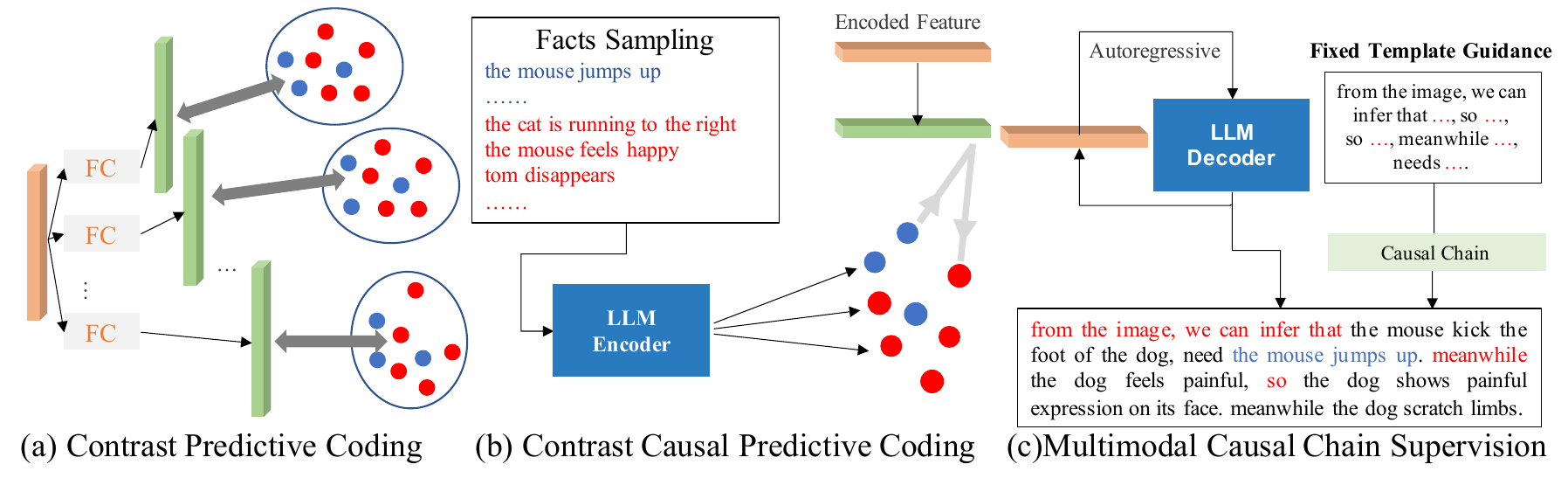}
	\caption{Structures of CPC (a), CCPC (b), and MCCS (c).}\label{fig8_10_cpc_mmcs}
\end{figure}

\subsection{Multimodal Causal Chain Supervision}\label{mmcs}

To enable better characterization of LLM-encoded features, introduce the supervision of the causal chain text generation.
Inspired by a work related to chain-of-thoughts \cite{kojima2022large,wei2022chain,zhang2022automatic,shum2023automatic,zhang2023multimodal}, we use text in the form of causal chains to provide supervised signals.
In the training phase, we generate text labels for samples that include causal chain annotations.
In particular, a fixed template guidance helps generate this part of the labels as shown in Figure \ref{fig8_10_cpc_mmcs}(c).
Specifically, these text labels provide an auxiliary optimization for the trainable parameter part of the model, as shown in Figure \ref{fig8_10_cpc_mmcs} (c).

\section{Experiments}\label{Experiments}

In this section, we show and analyze the experimental phenomena.
First, we compare the results of the three generative paradigms in Figure \ref{fig6_three_paradigm}.
After that, we block latent space translation (LST), CCPC, and multimodal causal chain supervision (MMCS) and analyze the results of the ablation experiments.
All experiments are run on an A100$\times$8 server.
In the dataset, we divided 17,524 samples into 15,524, 1,000 and 1,000, corresponding to the training, validation and testing sets.
Among them, 3809 samples in the training set include causal chain annotations.
Regarding the model, the LLM in this work references T5-XXL, and the image decoder uses stable diffusion.
In the training phase, we use Flan-T5-XXL \cite{raffel2020exploring,https://doi.org/10.48550/arxiv.2210.11416} with the original stable diffusion to initialize the parameters. All initial learning rates are set to 3e-5.
In the comparison experiments, we use ADAM \cite{kingma2014adam} as the optimizer. We set the batch size to 16 and the epoch to 20. 

Additional, we show more details and analysis in \textbf{\textit{Supplementary Material}}.

\subsection{Evaluation Metrics}\label{exp_em}

We design CLIP-based \cite{gal2022stylegan,radford2021learning} and human-based evaluation matrics, respectively.
Specifically, we compute the similarity of CLIP features between the generated image and the ground truth, denoted as $Sim_{Avg}$.
However, given the diversity that results from causal reasoning, it is not reasonable to conclude that a result different from GT is wrong.
Thus, we propose $Sim_{Best@k}$ to compute the maximum value of similarity among the $k$ results generated.
In our experiments, we set $k$ to 9.
After that, we introduce AUC based on CLIP score to observe the semantic accuracy of the generated pictures, denoted as $AUC_{Avg}$ and $AUC_{Best@k}$.
In addition, we incorporate human evaluations in order to accommodate the diversity of results.
We invite 10 algorithmic researchers to evaluate whether the generated images are semantically causally related to the input.
We invite 10 researchers to evaluate whether the generated images are semantically causally related to the input to obtain the accuracy.
In addition, we ask each evaluator to subjectively select the one they think is the best to compare the generative performance of the different methods, which is denoted as $Chosen Rate$.

\subsection{Causal Image Generation}\label{exp_cig}

\begin{figure*}[ht]
	\centering
	\includegraphics[width=0.88\linewidth]{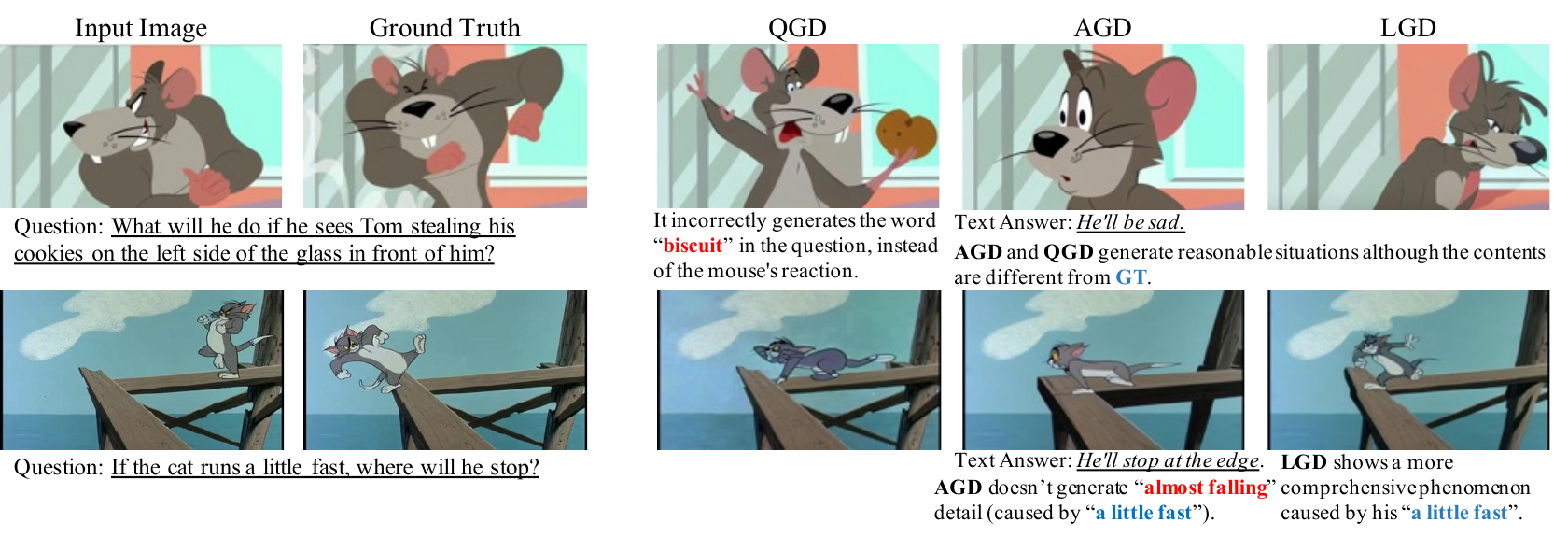}
	\caption{Visualization of the results generated by QGD, AGD, and LGD.}\label{fig11_exp_threeparadigm_short}
\end{figure*}

\begin{figure*}[ht]
	\centering
	\includegraphics[width=0.88\linewidth]{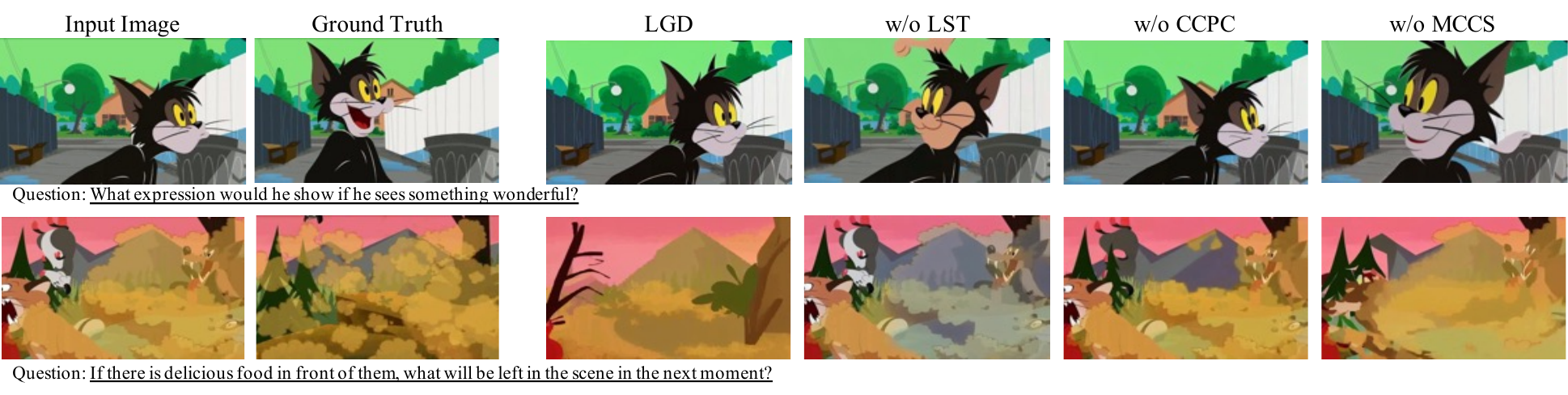}
	\caption{Visualization of the ablation results.}\label{fig12_exp_ablation_short}
\end{figure*}

We evaluate the three paradigms represented in Figure \ref{fig6_three_paradigm}, as shown in Figure \ref{fig11_exp_threeparadigm_short}.
It can be clearly observed that the results of the question-guided diffusion model (QGD) are confusing.
In particular, the image decoder may incorrectly add something from the question into the image rather than understanding the result to which this question would lead.
As shown in the first row of Figure \ref{fig11_exp_threeparadigm_short}, QGD incorrectly generates the content of the word ``biscuit'' instead of the rat's reaction to losing it.
This may be because the decoder does not have the ability to reason.
It can understand certain elements or variations that appear in the text and present them in the image modification process while ignoring those implicit inferences from the text.
More so, it makes this paradigm easily adaptable to tasks like image editing rather than causal content generation.
However, AGD is an improvement of QGD.
Since QGD lacks the ability of causal reasoning, a possible solution is to cascade a text reasoning model before the image decoder, whose duties are simultaneously reduced to a single editing model.
Fortunately, the examples in Figure \ref{fig11_exp_threeparadigm_short} show that AGD is effective.
Furthermore, The Latent-guided diffusion model (LGD) is a further improvement on the AGD.
This suggests that language does overlook some imperceptible variations in images that may be preserved in the hidden space, as shown in the second row of Figure \ref{fig11_exp_threeparadigm_short}.
In addition, we evaluate the three methods quantitatively, as shown in Table \ref{tab1_exp_threeparadigm}.
The results of all the experiments can be seen in Table 1, where LGD is superior in all metrics.

\begin{table}[!ht]
\scriptsize
\centering
\caption{{Quantitative comparison of three paradigms. CLIP and human evaluations of the mentioned methods.}\label{tab1_exp_threeparadigm}}
\tabcolsep=7pt
\setlength{\aboverulesep}{0pt}
\setlength{\belowrulesep}{0pt}
\begin{tabular}{l|c|c|c}
\toprule
\makebox[0.119\textwidth][c]{Methods}	&
\makebox[0.08\textwidth][c]{QGD}	&
\makebox[0.08\textwidth][c]{AGD}	&
\makebox[0.08\textwidth][c]{LGD}	\\
\midrule
{\makebox[0.119\textwidth][c]{$Sim_{Avg}$ (CLIP)}}
	&0.8361		&0.8444		&0.8589	\\
\midrule
{\makebox[0.119\textwidth][c]{$Sim_{Best@9}$ (CLIP)}}
	&0.8831		&0.8867		&0.9038	\\
\midrule
{\makebox[0.119\textwidth][c]{$AUC_{Avg}$ (CLIP)}}
	&0.8311		&0.8394		&0.8539	\\
\midrule
{\makebox[0.119\textwidth][c]{$AUC_{Best@9}$ (CLIP)}}
	&0.8781		&0.8819		&0.8987	\\
\midrule
{\makebox[0.119\textwidth][c]{$Acc$ (human)}}
	&0.1695		&0.1852		&0.3239	\\
\midrule
{\makebox[0.119\textwidth][c]{$Chosen Rate$ (human)}}
	&0.1601		&0.2310		&0.5135	\\
\toprule
\end{tabular}
\end{table}

\subsection{Ablation Study}\label{exp_as}

We conduct experiments to analyze the effects of three modules proposed before: latent space translation (LST), contrast causal predictive coding (CCPC), and multimodal causal chain supervision (MCCS). 
Figure \ref{fig12_exp_ablation_short} presents the experimental results obtained after removing these modules.
Precisely, the fourth, fifth, and sixth columns of Figure \ref{fig12_exp_ablation_short} correspond to the generated results when LST, CCPC, and MCCS are removed, respectively.
It can be observed that the absence of LST significantly degrades the quality of the generated images.
A possible reason is that LST effectively translates the output of the text encoder into features that the stable diffusion \cite{rombach2022high} understands, reducing the performance degradation caused by communication gaps.
Meanwhile, the lack of CCPC may cause the model to generate content that is more similar to the original image, supporting the notion that predictive coding is necessary for a task that requires generating content that has not yet occurred.
Additionally, the absence of MCCS leads to a higher likelihood of semantic errors in the generated content, which is reasonable considering that MCCS provides supervision for semantic understanding.

\section{Conclusion}\label{Conclusion}
In this study, we rethink image content generation and propose the task of causal image content generation. 
To support the task, we propose a dataset of VQAI based on the \textit{Tom and Jerry} cartoon series. 
Furthermore, we analyze the challenges of the task and propose an LGD approach, which is experimentally demonstrated in this paper. 
Finally, we further observe the experimental results of this task on several interesting aspects and analyze some of its potential and drawbacks.

\section{Acknowledgments}
This work was supported by the National Key R\&D Program of China (No. 2021ZD0113000) and the Innovative Development Joint Fund Key Projects of Shandong NSF (ZR2023LZH003).

\appendix
\section{Appendix I}
\section{VQAI Dataset Construction}\label{appendix_vqai}

This section presents some details about the dataset creation process, including the collection and review of samples, the composition of the dataset, and its distribution.

\subsection{Data Collection}\label{appendix_vqai_collect}

The whole collection process is divided into five steps.
First, we collect suitable videos.
Then, we divide these videos into several segments according to certain rules.
Next, we hire annotators to extract relevant image pairs from these segments.
Finally, we instruct annotators to label appropriate causal questions for these image pairs.
As an additional step, we annotate causal chains for a subset of the samples.

\noindent{\textbf{\textit{Video Collection:}}}
We create the dataset based on the \emph{Tom and Jerry} cartoon series, as it greatly minimizes the use of language and has a more straightforward worldview compared to natural scenes.
We collect a selection of \emph{Tom and Jerry} cartoons released between 1940 and 2021.
These videos mostly have a duration of around 6-7 minutes, with a few, such as those from \emph{The Tom and Jerry Show (Season 1)}, exceeding 10 minutes in length.
Finally, we collect a total of 755 episodes from public sources.

\noindent{\textbf{\textit{Video Segmentation:}}}
To facilitate the extraction of image pairs capturing causal variations from the videos, we automatically segment each video into smaller segments, with each segment containing a single storyline. 
We employe pixel difference calculations between adjacent frames to determine whether they belong to the same scene.
Specifically, we set thresholds for the pixel differences and minimum segment lengths to ensure the coherence of each segment.
This approach is instrumental in improving the efficiency of the annotators' work.

We use the open-source software \emph{SceneDetect}\footnotemark[1] for the segmentation process.
On average, each episode is divided into approximately 35 segments.
The introductory and concluding segments are manually removed, and the remaining segments are assigned to the annotators.
\footnotetext[1]{\url{https://github.com/Breakthrough/PySceneDetect/tree/master/scenedetect}}

\noindent{\textbf{\textit{Video Screenshot:}}}
The annotators are instructed to extract a pair of images from each segment that exhibit a causal relationship.
Typically, the second image depicts a content that is highly likely to occur based on the condition depicted in the first image.
Considering that this work describes an image generation task involving implicit conditions, it is emphasized that the second image, which serves as the ground truth, must be generatable based on the given one.
In other words, the difference between the two images should not be too large, and the difference should not be unpredictable.
Therefore, we require the annotators to carefully assess the variations between the two screenshots.
Specifically, we require the captured screenshots to meet the following rules:

\begin{itemize}
\item{The camera's shooting angle remains constant, and there is little to no shaking, displacement, zooming, or panning between the two captured images.}
\item{Only a few specific characters or objects undergo changes in their shape, emotion, disappearance, or other state changes.}
\item{There is no change or only minor changes in all objects or characters in the non-prominent background.}
\item{There is no change in global information such as brightness, color, etc., except for samples that specifically inquire about scene variations.}
\end{itemize}

\noindent{\textbf{\textit{Question Annotation:}}}
We hire several annotators to annotate the questions for the image pairs mentioned above.
A valid VQAI sample should satisfy the following criteria: the change between the second image and the first one reasonably answers the given question, and the second image can be derived by simple logical reasoning by humans.
Therefore, to ensure the quality of question annotations, we implement a whitelist approach to constrain the work of annotators.
We predefine five categories of sample formats that meet the requirements, and any samples that cannot be classified into these categories will be rejected.

The requirements for the five categories of samples are as follows:

\begin{figure*}[ht]
	\centering
	\includegraphics[width=0.9\linewidth]{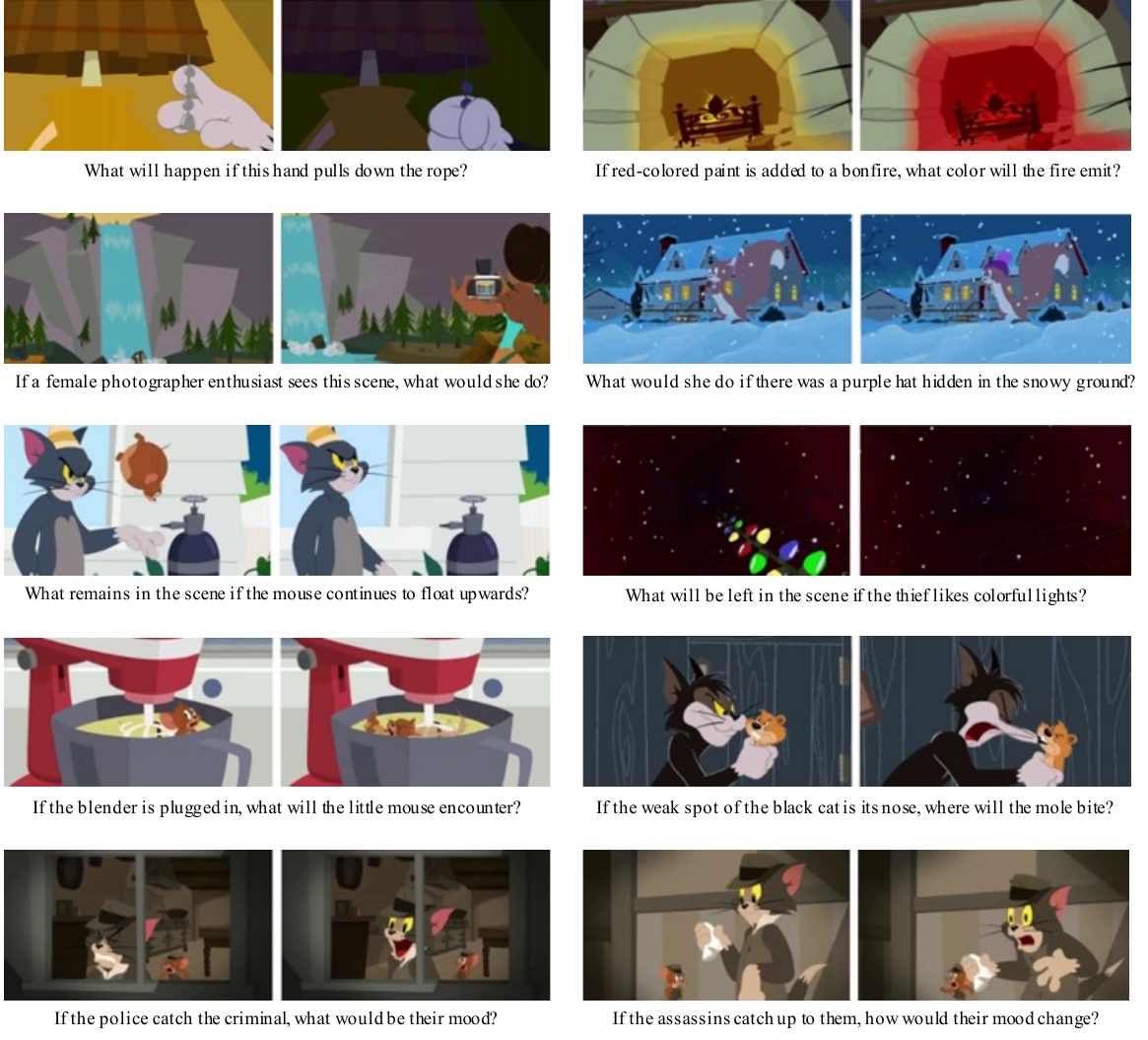}
	\caption{More examples of the five categories of samples are provided in the following order, from top to bottom: Scenery Variation, More Entities, Fewer Entities, Entities Variation, and Emotion Variation.}\label{fig16_demo_5_cates}
\end{figure*}

\textbf{Scenery Variation:}
The scene or environment is modified.

The scene changes may include variations in brightness, season, weather, temperature, color, and other related factors.
However, it is essential to ensure that the two images have consistent camera angles.
Additionally, except for the mentioned aspects, the changes in other objects and characters should be minimal unless there is a clear cause-effect relationship with the scene change.
For example, ``The mouse shivers due to the weather changing to snowfall.''

Under such circumstances, the textual annotations should focus on the \textbf{conditions} under which the change occurs.
For example, for the scenario of ``transition from night to day,'' you can phrase the question as ``What will this place look like in another 12 hours?'' instead of ``What will it look like when it's daylight?''

\textbf{More Entities:}
One or more entities have been added.

The changes in this type of sample include the addition of one or more characters or objects.
You need to ensure that apart from the addition, there are no other changes in the two images, unless the change is a result of the presence of that entity, for example, ``the appearance of \emph{Tom}'' causing ``\emph{Jerry} to become frightened.''

For this type of sample, the question should provide the necessary conditions that led to the addition of the new entity.
When the entity is not commonly seen, it's important to give specific hints about its size, shape, or other attributes.
For example, if a blue tent is added to the scene, you should express in the question that it is ``blue'' to provide accurate supervision signals for our generation model.

\textbf{Fewer Entities:}
One or more entities have been reduced.

The change in this type of sample is that one or more entities are either in the process of or have already exited the scene.
Other changes in the scene should be minimal unless they are caused by the disappearance of the target, such as ``the mouse floating upwards out of the frame'' causing ``the cat looking up''.

For this type of sample, the questions should be specific and explicitly include the cause of the target's disappearance.
If necessary, additional conditions can be added to establish a direct causal relationship between the change and the question.

\textbf{Entities Variation:}
Changes occur in entities or relationships between them.

In this type of sample, there is no change in the camera angle, the number of objects or characters remains the same, and the change only occurs within existing entities or their relationships.
This is the most common type of sample, where typically only two or fewer characters or objects undergo a change.
These samples are excellent for expressing causality.
Usually, the change is explicit and it is easy to annotate a reasonable question.

During annotating, it's important to ensure that the given condition in the question can be fully inferred to explain the change.
If the change, such as ``the mole bites the black cat's nose,'' cannot be deduced directly from the image or through common sense, it's necessary to provide additional hints or expressions, such as ``the black cat's weakness is its nose,'' to clarify the context and help understand the change.

\textbf{Emotion Variation:}
One or more characters' emotions change, accompanied by facial expressions or movements being modified.

This type of variation is also very common in videos because the emotions of all characters are explicit, and almost all plots are accompanied by changes in different character emotions.
Usually, images focus on close-ups of certain characters, so there are rarely significant changes in the frame.
However, you need to be cautious that the number of such samples should not be too high (lower than 20\%) because although they are easy to annotate, they are relatively easy for our model.

Similarly, when annotating questions, you should not directly ask what emotion it is, but rather provide a specific event that caused the emotion variation.

Additionally, all questions should focus on the changes between the two images, regardless of the category to which the change belongs.
The question should provide a reasonable condition that leads to the change without directly revealing it.
We provide more sample examples in Figure \ref{fig16_demo_5_cates}, arranged from top to bottom corresponding to the whitelist for the five categories of question annotations.

\noindent{\textbf{\textit{Causal Chain Annotation:}}}
The causal chains are annotated out by three experienced researchers, and each node in the causal chain is represented by a binary tuple of entity and its variation.
This part of the work is done based on the annotated casual questions.
In other words, the questions can be considered as the starting nodes in the causal chains.

\subsection{Data Statistics}\label{appendix_vqai_statistic}

This section analyzed the components and attributes of VQAI, including the distribution statistics of different sample categories, the length statistics of questions and causal chains annotations.

\begin{figure}[ht]
	\centering
	\includegraphics[width=1.0\linewidth]{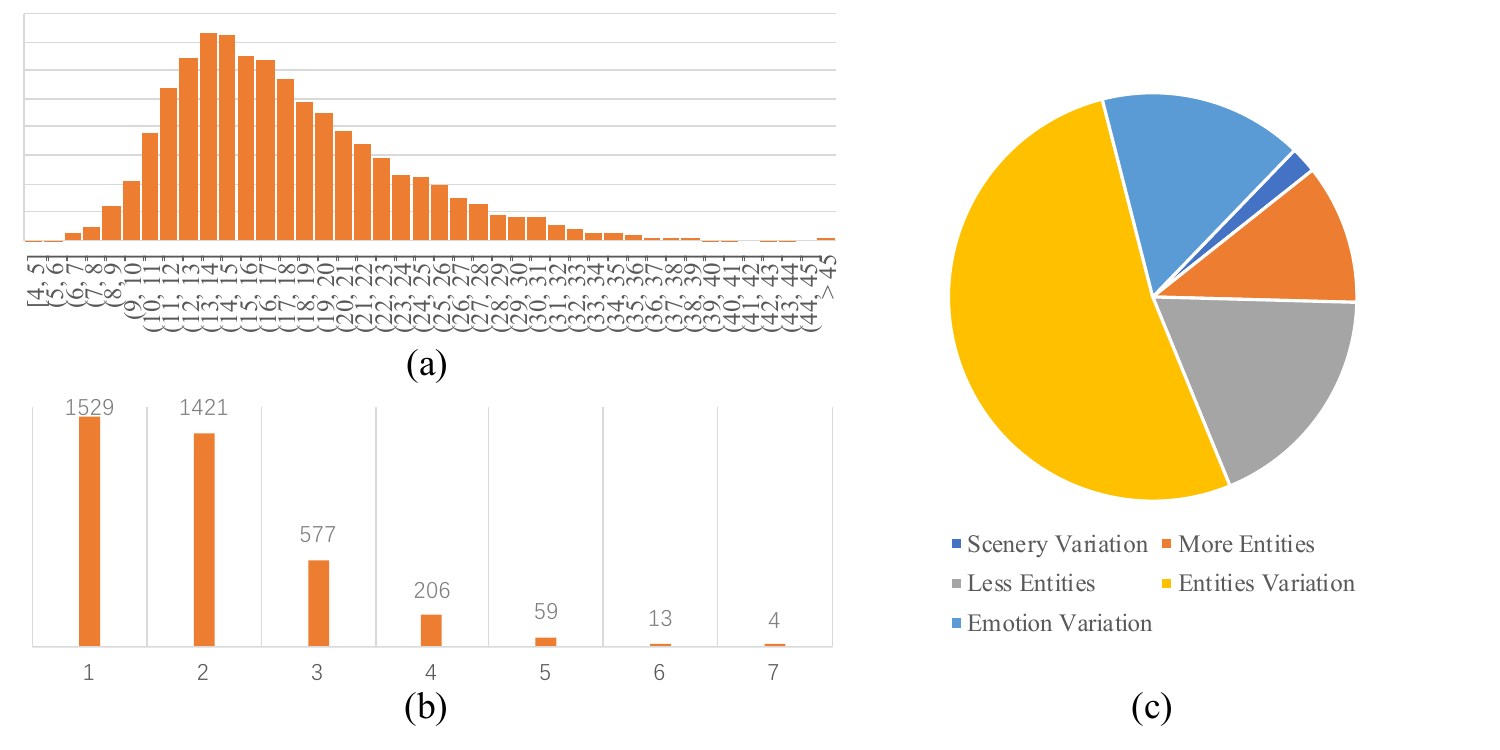}
	\caption{Data statistics. From (a) to (c) are the statistics of question lengths, causal chain lengths, and sample category distribution.}\label{fig18_distributions}
\end{figure}

Figure \ref{fig18_distributions} (a) shows the distribution of question lengths for the annotated questions, and it can be observed that the majority of questions are concentrated between 10 and 25 words.
The average length of these questions is approximately 18.1 words.
Furthermore, it is evident that a small portion of questions exceeds 30 words, indicating that certain image changes are difficult to be simply described by conditions, prompting annotators to provide additional hints.

Figure \ref{fig18_distributions} (b) shows the claimed long-tail effect of causal chain annotations mentioned in the main text.
It is evident that the majority of causal chains consist of only 1-2 nodes, indicating that most image inferences can be accomplished within 1-2 steps.
For samples with longer chains, they typically exhibit more complex relationships, requiring the model to generate multiple variations.

Figure \ref{fig18_distributions} (c) displays the distribution of the five types of samples mentioned earlier.
Obviously, samples related to scene variations are the least common, while samples involving entity variations are the most abundant, aligning with our intuition about the collected videos.
It is worth noting that samples related to emotions have been compressed to approximately 16.1\% of the overall distribution, indicating that our reminders and reviews for the annotators have been effective.
We hope that these actions contribute to making the dataset more compelling.

However, our dataset still has some limitations that can serve as motivation for future improvements.
For instance, the current distribution of the dataset remains highly imbalanced, making it more challenging to evaluate models effectively.
Furthermore, using \emph{Tom and Jerry} cartoons as the data source is indeed a bold simplification of the causal content generation problem.
Therefore, exploring causal generation problems in the context of the natural world would be an interesting research direction to consider.

\subsection{Legal Sourcing and Intended Usage}\label{appendix_vqai_legal}

All the videos used in the collection of VQAI are sourced from the \emph{Tom and Jerry} animated series, which is copyrighted material.
Hence, we follow Fair Use §107: ``the fair use of a copyrighted work, including such use by $\cdots$ scholarship, or research, is not an infringement of copyright'', where fair use is determined by ``the purpose and character of the use, including whether such use is of a commercial nature or is for nonprofit educational purposes'', ``the amount and substantiality of the portion used in relation to the copyrighted work as a whole'', and ``the effect of the use upon the potential market for or value of the copyrighted work.''
Hence, the VQAI datasets are noncommercial and should only be used for the purpose of academic research.

We bear all responsibility in case of violation of rights. 
The videos used during data collection are be under copyright, so we do not provide an official license and rely on Fair Use §107.
Our annotations are available under a CC BY-SA 4.0 license.


\section{Appendix II}
\section{Additional Experiments and Discussions}\label{appendix_exp}

\begin{figure}[ht]
	\centering
	\includegraphics[width=0.9\linewidth]{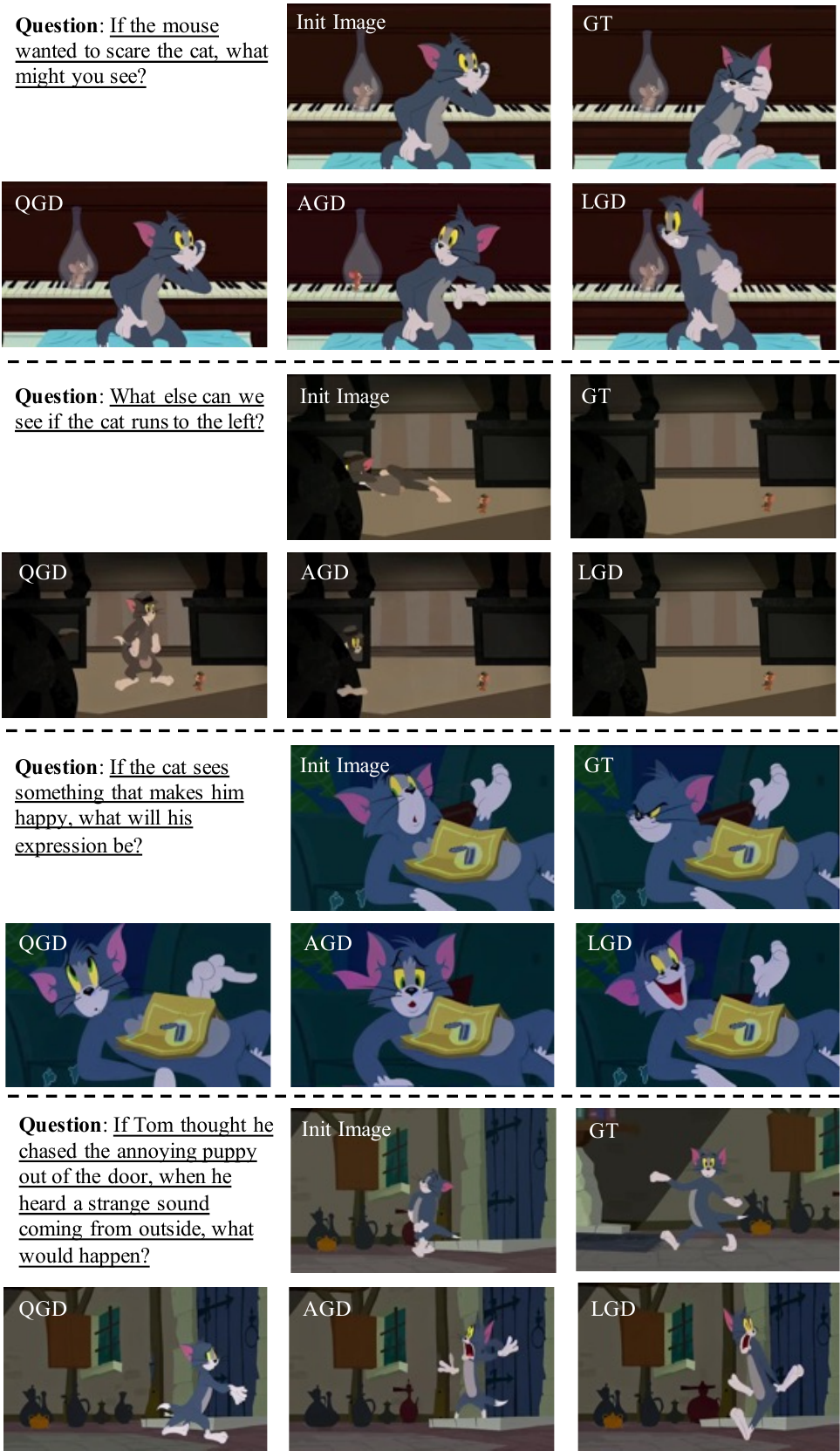}
	\caption{More visualization of the results generated by the proposed methods. Each sample is presented in turn with the \textit{Question}, the \textit{Init Image}, its \textit{ground truth}, and the results of \textit{QGD}, \textit{AGD}, and \textit{LGD}.}\label{fig11_exp_threeparadigm}
\end{figure}

\begin{figure}[ht]
	\centering
	\includegraphics[width=0.9\linewidth]{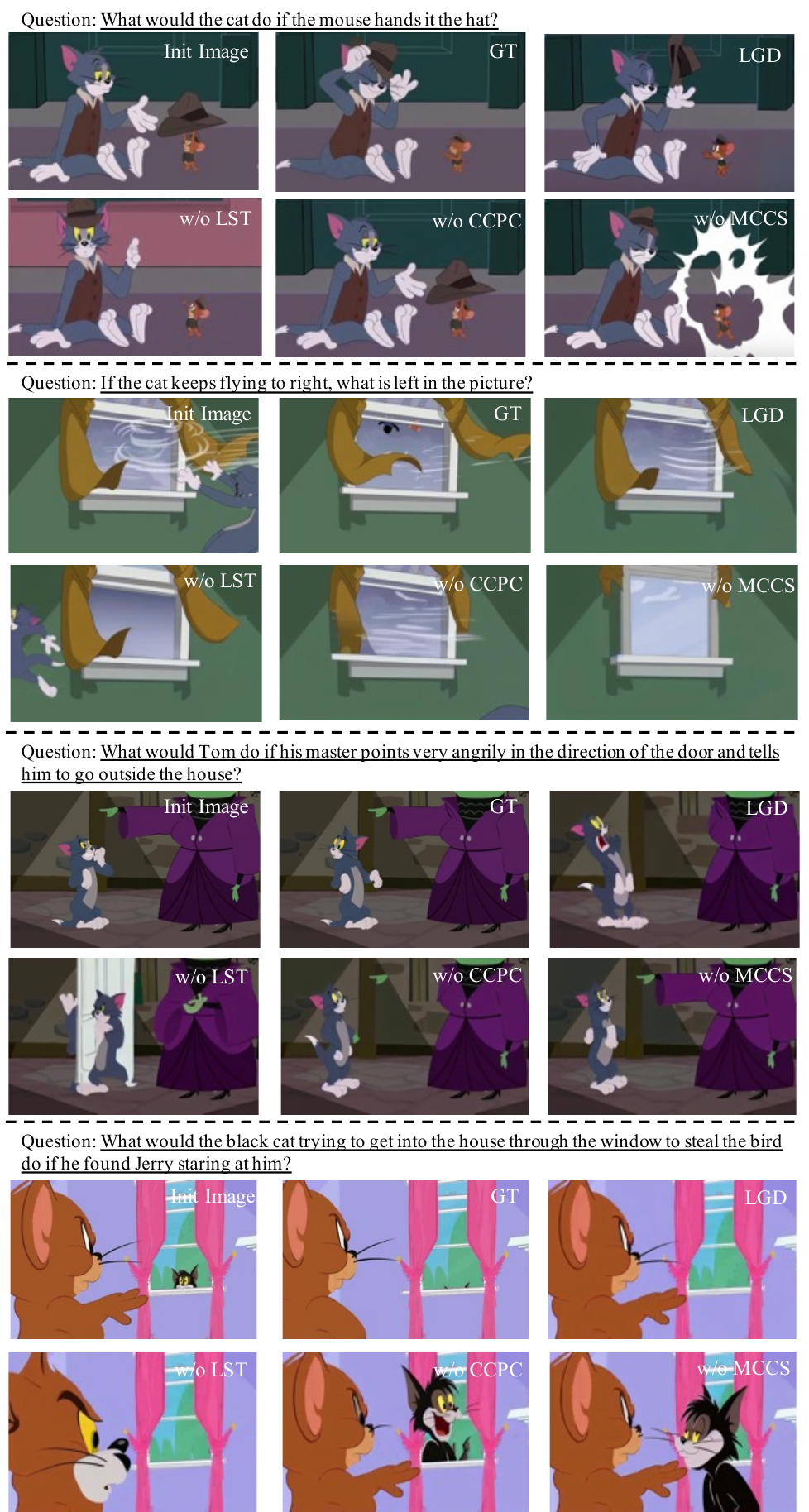}
	\caption{Visualization of the ablation results. Each sample is presented in turn with the \textit{Question}, the \textit{Init Image}, its \textit{ground truth}, the result of \textit{LGD}, and the results generated without \textit{LST}, \textit{CCPC}, and \textit{MCCS}.}\label{fig12_exp_ablation}
\end{figure}

\subsection{Supplements for Evaluation Metrics}

As discussed in Section \textit{Experiments}, we evaluate the performance of different generated images in terms of accuracy, diversity based on CLIP \cite{gal2022stylegan,radford2021learning} and human, respectively. 
For CLIP-related metrics, we load ViT-B/32, whose weights are trained and publicly available\footnotemark[2], as the encoder, and evaluate it based on our adaptation of an open-source code\footnotemark[3].
All features used to compute similarity are 512 dimensional.
We supplement here a detailed description of the metrics that appear in Table \ref{tab1_exp_qgd_5c}-\ref{tab1_exp_lgd_5c}.
$Sim_{Avg}$ is used to represent the average cosine similarity between the generated images and the ground truth, as given in Function \ref{formula1}, where $Cos(*)$ denotes the computation of cosine similarity. 
\footnotetext[2]{\url{https://huggingface.co/sentence-transformers/clip-ViT-B-32}}
\footnotetext[3]{\url{https://github.com/Taited/clip-score}}
\begin{equation}
\\Sim_{Avg} = \frac{{\sum_{i=1}^{N}\sum_{k=1}^{K}{Cos(f_{pred_{i,k}}, f_{gt_i})}}}{{N \times K}}
\label{formula1}
\end{equation}
\begin{equation}
\\Sim_{Best@K} = \frac{{\sum Max({Cos(f_{pred_{i,k}}, f_{gt_i})}|k \in [1, K])}}{{N}}
\label{formula2}
\end{equation}
However, it assumes by default that the ground truth is the only correct answer, which is inconsistent with the fact that there is diversity in the results.
To alleviate this conflict, we designed $Sim_{Best@K}$, with the formula shown in Function \ref{formula2}.
Whenever one of several predictions hits the ground truth, the model gets a high score.
Meanwhile, we test the accuracy of the generated images under different thresholds of CLIP similarity and evaluate the AUC of this accuracy curve, as shown in Table \ref{tab1_exp_qgd_5c}-\ref{tab1_exp_lgd_5c}.
Particularly, we compute the average and maximum scores separately for each sample, denoted as $AUC_{Avg}$ and $AUC_{Best@K}$, respectively.
In all experiments in this study, $K$ is set to 9.

However, as described in the main text, these CLIP-based evaluation metrics still lack the ability to assess the diversity of the generated images.
Thus, a metric that can assess the correctness of each generated image is necessary.
Unfortunately, current automated assessment methods cannot do this because there is only one ground truth.
As a result, we introduce person evaluation metrics.
We invite 10 algorithm developers to participate in the evaluation of the generated results.
Images generated by multiple methods are randomly mixed together, and each evaluator is asked to select all semantically plausible ones from them.
We calculate the correctness of the content generated by each method in this way, denoted as $Acc$.
In addition, to assess which method is the best, each evaluator also needs to pick a best image among all correct predictions.
Specifically, we compare the performance of the different methods more intuitively by counting the best of these.
We refer to this metric as $Chosen Rate$.

Furthermore, to complement the assessment of image quality, we additionally supplement this section (in \textit{Supplementary Material}) with an evaluation of Fréchet inception distance (FID) score \cite{yu2021frechet}, as shown in Table \ref{tab1_exp_qgd_5c}-\ref{tab1_exp_lgd_5c}.

\subsection{More experimental results}

\subsubsection{More experiments on Causal Image Generation}\

We present additional experimental results in Figure \ref{fig11_exp_threeparadigm} to compare the performance of QGD, AGD, and LGD.
The experimental results demonstrate that QGD struggles to generate results requiring reasoning, while AGD can realize it.
For example, in the first case shown in Figure \ref{fig11_exp_threeparadigm}, QGD does not generate a reasonable result.
This may be because QGD fails to understand what the ``mouse trying to scare the cat'' thing means, and likewise fails to understand what it might lead to.
In the AGD results, at least the cat shows reasonable feedback.
Although it doesn't exhibit panic-like emotions, like Ground Truth, ``calmly looking back'' is also a possible response from the cat.

\begin{table}[!ht]
\scriptsize
\centering
\caption{{Quantitative results of QGD on each type of sample. From left to right, denote \textit{Total Testset}, \textit{Scenery Variation}, \textit{More Entities}, \textit{Fewer Entities}, \textit{Entities Variation}, and \textit{EMotion Variation}, respectively.}\label{tab1_exp_qgd_5c}}
\tabcolsep=7pt
\setlength{\aboverulesep}{0pt}
\setlength{\belowrulesep}{0pt}
\begin{tabular}{l|c|c|c|c|c|c}
\toprule
\makebox[0.1\textwidth][c]{Methods}	&
\makebox[0.02\textwidth][c]{TT}	&
\makebox[0.02\textwidth][c]{SV}	&
\makebox[0.02\textwidth][c]{ME}	&
\makebox[0.02\textwidth][c]{FE}	&
\makebox[0.02\textwidth][c]{EV}	&
\makebox[0.02\textwidth][c]{EMV}	\\
\midrule
{\makebox[0.119\textwidth][c]{$Sim_{Avg}$ (CLIP)}}
	&0.836		&0.762		&0.826		&0.797		&0.845		&0.870	\\
\midrule
{\makebox[0.119\textwidth][c]{$Sim_{Best@9}$ (CLIP)}}
	&0.883		&0.814		&0.872		&0.856		&0.889		&0.911	\\
\midrule
{\makebox[0.119\textwidth][c]{$AUC_{Avg}$ (CLIP)}}
	&0.831		&0.757		&0.821		&0.792		&0.839		&0.865	\\
\midrule
{\makebox[0.119\textwidth][c]{$AUC_{Best@9}$ (CLIP)}}
	&0.878		&0.808		&0.867		&0.851		&0.883		&0.907	\\
\midrule
{\makebox[0.119\textwidth][c]{$Acc$ (human)}}
	&0.169		&0.02		&0.197		&0.106		&0.176		&0.216	\\
\midrule
{\makebox[0.119\textwidth][c]{$Chosen Rate$ (human)}}
	&0.160		&0.0			&0.187		&0.121		&0.168		&0.179	\\
\midrule
{\makebox[0.119\textwidth][c]{FID$\downarrow$}}
	&38.0		&277.0		&126.7		&119.3		&58.2		&88.8	\\
\toprule
\end{tabular}
\end{table}

\begin{table}[!ht]
\scriptsize
\centering
\caption{{Quantitative results of AGD on each type of sample.}\label{tab1_exp_agd_5c}}
\tabcolsep=7pt
\setlength{\aboverulesep}{0pt}
\setlength{\belowrulesep}{0pt}
\begin{tabular}{l|c|c|c|c|c|c}
\toprule
\makebox[0.1\textwidth][c]{Methods}	&
\makebox[0.02\textwidth][c]{TT}	&
\makebox[0.02\textwidth][c]{SV}	&
\makebox[0.02\textwidth][c]{ME}	&
\makebox[0.02\textwidth][c]{FE}	&
\makebox[0.02\textwidth][c]{EV}	&
\makebox[0.02\textwidth][c]{EMV}	\\
\midrule
{\makebox[0.119\textwidth][c]{$Sim_{Avg}$ (CLIP)}}
	&0.844		&0.785		&0.831		&0.804		&0.853		&0.878	\\
\midrule
{\makebox[0.119\textwidth][c]{$Sim_{Best@9}$ (CLIP)}}
	&0.887		&0.825		&0.876		&0.856		&0.893		&0.916	\\
\midrule
{\makebox[0.119\textwidth][c]{$AUC_{Avg}$ (CLIP)}}
	&0.839		&0.779		&0.827		&0.800		&0.849		&0.874	\\
\midrule
{\makebox[0.119\textwidth][c]{$AUC_{Best@9}$ (CLIP)}}
	&0.882		&0.820		&0.871		&0.851		&0.889		&0.911	\\
\midrule
{\makebox[0.119\textwidth][c]{$Acc$ (human)}}
	&0.185		&0.131		&0.215		&0.149		&0.177		&0.232	\\
\midrule
{\makebox[0.119\textwidth][c]{$Chosen Rate$ (human)}}
	&0.231		&0.161		&0.187		&0.223		&0.191		&0.204	\\
\midrule
{\makebox[0.119\textwidth][c]{FID$\downarrow$}}
	&34.4		&261.1		&125.0		&115.6		&54.7		&87.8	\\
\toprule
\end{tabular}
\end{table}

\begin{table}[!ht]
\scriptsize
\centering
\caption{{Quantitative results of LGD on each type of sample.}\label{tab1_exp_lgd_5c}}
\tabcolsep=7pt
\setlength{\aboverulesep}{0pt}
\setlength{\belowrulesep}{0pt}
\begin{tabular}{l|c|c|c|c|c|c}
\toprule
\makebox[0.1\textwidth][c]{Methods}	&
\makebox[0.02\textwidth][c]{TT}	&
\makebox[0.02\textwidth][c]{SV}	&
\makebox[0.02\textwidth][c]{ME}	&
\makebox[0.02\textwidth][c]{FE}	&
\makebox[0.02\textwidth][c]{EV}	&
\makebox[0.02\textwidth][c]{EMV}	\\
\midrule
{\makebox[0.119\textwidth][c]{$Sim_{Avg}$ (CLIP)}}
	&0.859		&0.812		&0.835		&0.836		&0.865		&0.888	\\
\midrule
{\makebox[0.119\textwidth][c]{$Sim_{Best@9}$ (CLIP)}}
	&0.904		&0.856		&0.883		&0.893		&0.908		&0.923	\\
\midrule
{\makebox[0.119\textwidth][c]{$AUC_{Avg}$ (CLIP)}}
	&0.854		&0.807		&0.830		&0.831		&0.860		&0.883	\\
\midrule
{\makebox[0.119\textwidth][c]{$AUC_{Best@9}$ (CLIP)}}
	&0.899		&0.851		&0.878		&0.889		&0.903		&0.918	\\
\midrule
{\makebox[0.119\textwidth][c]{$Acc$ (human)}}
	&0.324		&0.351		&0.309		&0.272		&0.276		&0.494	\\
\midrule
{\makebox[0.119\textwidth][c]{$Chosen Rate$ (human)}}
	&0.514		&0.833		&0.468		&0.545		&0.494		&0.522	\\
\midrule
{\makebox[0.119\textwidth][c]{FID$\downarrow$}}
	&32.4		&267.6		&128.5		&109.1		&53.4		&86.2	\\
\toprule
\end{tabular}
\end{table}

Meanwhile, LGD can generate more specific visual content compared to AGD.
For example, in the last case, AGD can reasonably predict that ``Tom'' will ``jump up in fright'' and ``show a frightened expression'' after ``hearing a strange noise outside the door''.
However, LGD presents more precise additional information in addition to this.
Since the ``sound'' comes from outside the door, ``Tom'' will most likely ``look back at the door''.
These details are likely to be missed due to the absence of the textual answer in the AGD.

Besides, Table \ref{tab1_exp_qgd_5c}-\ref{tab1_exp_lgd_5c} also shows the quantitative evaluation of the samples in each category.
We also demonstrate the FID score in Table \ref{tab1_exp_qgd_5c}-\ref{tab1_exp_lgd_5c} for assessing quality of the generated images.
The experimental results show that the samples belonging to the ``Emotion Variation'' category are the easiest to predict, probably because the degree of variation is usually small, as shown in Figure \ref{fig16_demo_5_cates}.
Conversely, ``Scenery Variation'' is the worst, probably because this type of sample is usually accompanied by changes in the whole image, resulting in a relatively poor score.

In addition, our quantitative assessment reveals some interesting phenomena.
For example, in human evaluations of QGD in scenery-variation samples, the accuracy is zero.
This means that it is very difficult for a QGD to understand how to modify a scenario when given only a description of the premise rather than a definitive textual guidance.
The FID scores illustrate that LGD not only improves accuracy, but likewise improves the quality of the generated images by a small margin.

\begin{table}[!ht]
\scriptsize
\centering
\caption{{Quantitative demonstration of ablation experiments: results without \textit{LST}, \textit{CCPC} and \textit{MCCS}, respectively.}\label{tab2_exp_ablation}}
\tabcolsep=7pt
\setlength{\aboverulesep}{0pt}
\setlength{\belowrulesep}{0pt}
\begin{tabular}{l|c|c|c|c}
\toprule
\makebox[0.119\textwidth][c]{Methods}	&
\makebox[0.06\textwidth][c]{LGD}	&
\makebox[0.06\textwidth][c]{w/o LST}	&
\makebox[0.06\textwidth][c]{w/o CCPC}	&
\makebox[0.06\textwidth][c]{w/o MCCS}	\\
\midrule
{\makebox[0.119\textwidth][c]{$Sim_{Avg}$ (CLIP)}}
	&0.8589		&0.8414		&0.8482		&0.8487\\
\midrule
{\makebox[0.119\textwidth][c]{$Sim_{Best@9}$ (CLIP)}}
	&0.9038		&0.8901		&0.9027		&0.9028\\
\midrule
{\makebox[0.119\textwidth][c]{$AUC_{Avg}$ (CLIP)}}
	&0.8539		&0.8361		&0.8432		&0.8433\\
\midrule
{\makebox[0.119\textwidth][c]{$AUC_{Best@9}$ (CLIP)}}
	&0.8987		&0.8981		&0.8986		&0.8987\\
\toprule
\end{tabular}
\end{table}

\subsubsection{More experiments on Ablation Study}\

Figure \ref{fig12_exp_ablation} demonstrates the effect of the lack of LST.
It is observed that the overall quality of the generated images decreases, i.e., the face of \textit{Jerry} in the last case.
A plausible reason for this is the conflict of text encoders.
Particularly, existing open-source weights for stable diffusion are usually based on CLIP \cite{radford2021learning}, which may not be as good as other LLMs regarding semantic understanding.
In this work, we use T5 \cite{raffel2020exploring} as the LLM, which makes LST look more like an adapter between the diffusion model and T5.
This may ease the difficulty of training all the cross-attention layers in its network, making the experimental results look much better.
The 3rd column of Table \ref{tab2_exp_ablation} indicates the practical effect without LST.
It might take far more samples to train to compensate for the effect of LST.

From the first three cases in Figure \ref{fig12_exp_ablation}, we can easily see that LGD without CCPC can easily generate results that are very similar to the initial image.
This indicates that CCPC somewhat pushes the features in the hidden space into the prediction space. 
However, from the perspective of quantitative evaluation, the gain is not obvious.

Meanwhile, the visualization in Figure \ref{fig12_exp_ablation} illustrates that LGD without MCCS can also generate semantically reasonable results.
However, the lack of MCCS may lead to some excessive changes.
Although these changes are also acceptable to a certain extent, this leads to poorer generation scores.

\subsection{More demonstrations and discussions}

\begin{figure}[ht]
	\centering
	\includegraphics[width=0.95\linewidth]{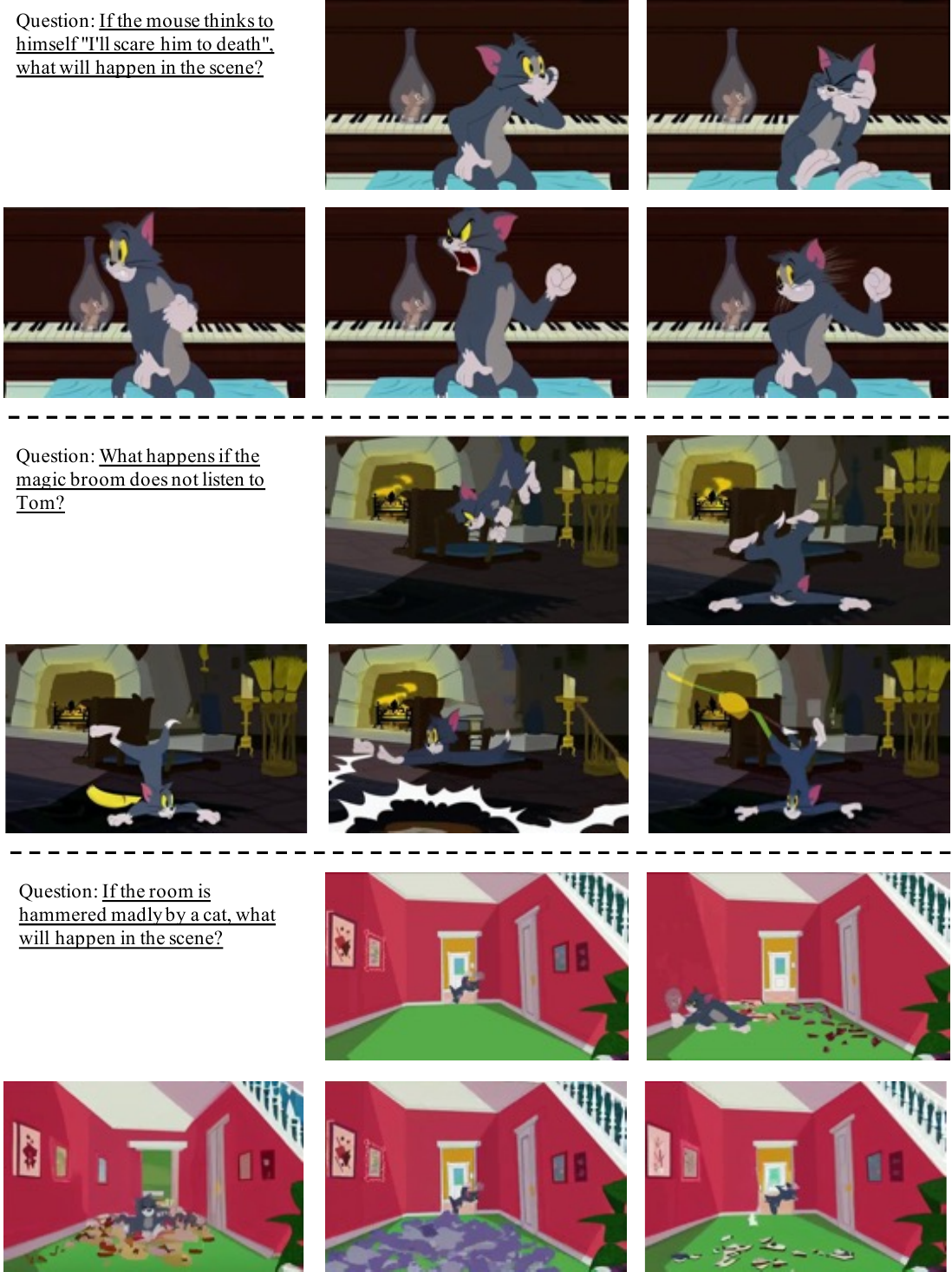}
	\caption{Visualization of generations initialized with various seeds. For each sample, the first row represents its initial image and ground truth, and the second row represents its generation results initialized via three different seeds.}\label{fig13_exp_seeds}
\end{figure}

\begin{figure}[ht]
	\centering
	\includegraphics[width=0.85\linewidth]{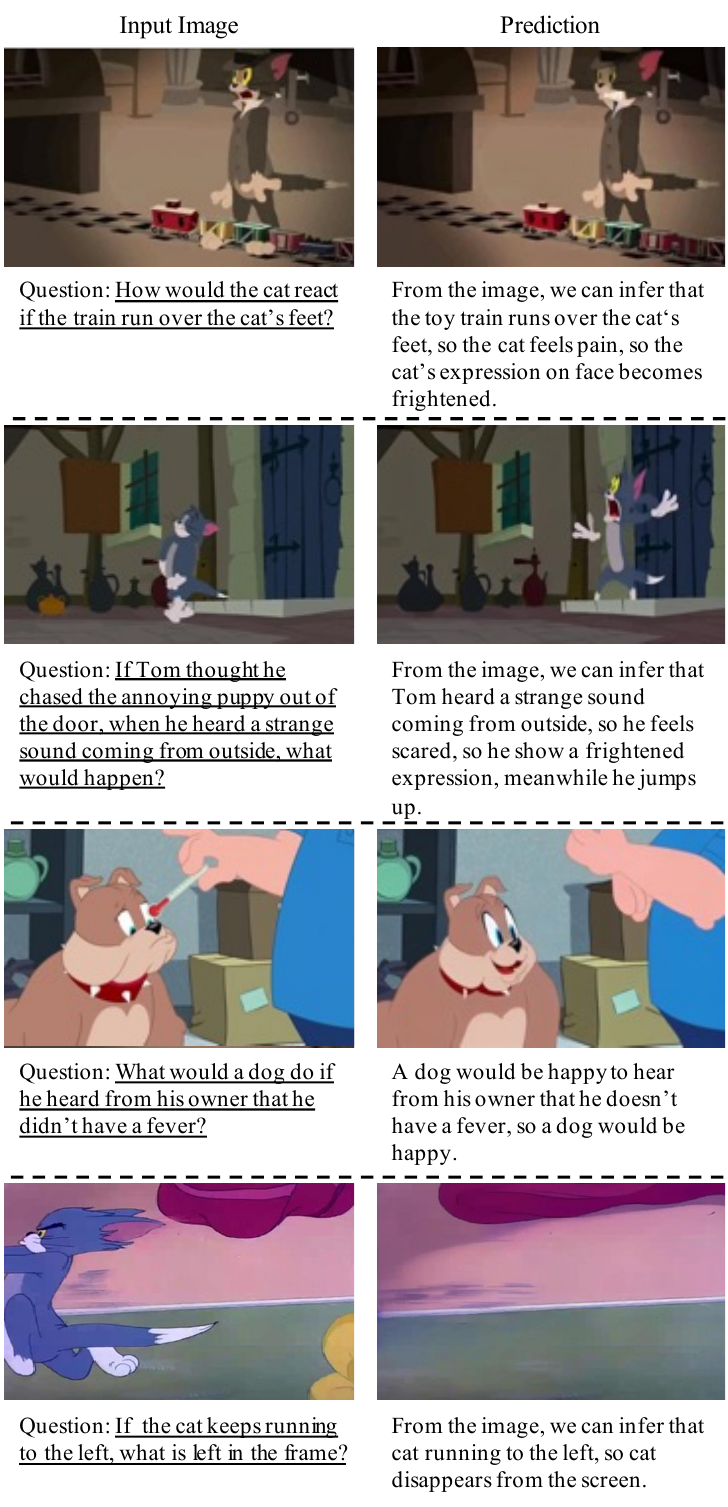}
	\caption{Visualization of the generated multimodal causal chains.}\label{fig14_exp_mcc}
\end{figure}

\subsubsection{Discussion on the diversity of generated content}\

Figure \ref{fig13_exp_seeds} demonstrates the diversity of the generated results.
Interestingly, as we set different seeds, we are likely to get multiple reasonable predictions.
As an illustration, in the 1st case, ``Tom'' shows emotions such as ``panic'', ``anger'', and ``fear'' under the same conditions.
Although these are different from ground truth, it is hard to say that they are unreasonable.
In the 2nd case, we can infer that ``Tom may fall on the ground'' according to the condition, and the different seeds lead to different postures of ``Tom'', which are all consistent with the semantic meaning of ``fall''.
The remaining two samples also show reasonable predictions of the generative model with different seeds.
Previously, we had been concerned that the model would overlearn fixed causal collocations and fall into the trap of pattern collapse.
Fortunately, these experimental results demonstrate the diversity of causal content generation.

\subsubsection{Demonstrations of generated results for Multimodal Causal Chain}\

Furthermore, we show some text generation results for multimodal causal chains in Figure \ref{fig14_exp_mcc} and image editing results based on these texts.
For example, in the 1st case, the text results generate information such as ``the cat feels pain'', thus ``the cat's expression becomes frightened" in an orderly manner.

However, these results also exhibit some shortcomings.
For example, in the 2nd case, a more detailed description is not given (compared to the LGD results in Figure \ref{fig11_exp_threeparadigm_short}).
Additionally, in the prediction on the top right, some repeated words are mentioned several times.
Despite these drawbacks, we believe that the LLM-based multimodal causal chain generation task is still a very interesting topic, considering that these results were only fine-tuned with less than 4000 training samples.

\begin{figure}[ht]
	\centering
	\includegraphics[width=0.95\linewidth]{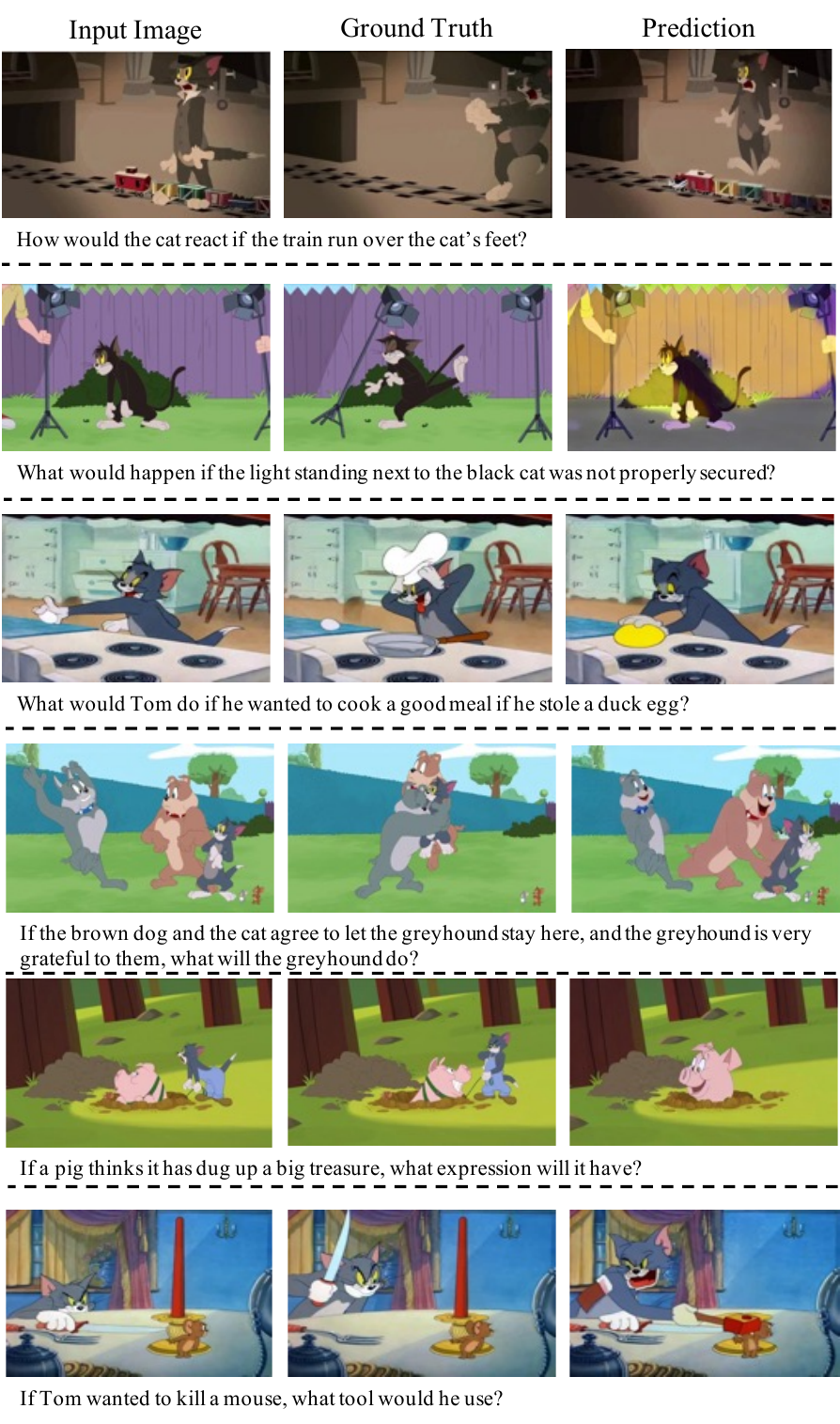}
	\caption{Visualization of some bad cases.}\label{fig15_exp_bad}
\end{figure}

\subsubsection{Experimental results of LGD with text answer}\

Given the phenomena manifested in the generated results of the multimodal causal chain, we consider concatenating the generated text into LGD.
The LGD{+} in Table \ref{tab3_exp_lgdplus} demonstrates such concatenation's effect and better performance in the 1,000 test samples.
It is intuitively reasonable that the experimental results make sense, considering that the text must provide a more direct semantic complement to the hidden space features.

\begin{table}[!ht]
\scriptsize
\centering
\caption{{Comparison of the results whether text answers are included in LGD or not.}\label{tab3_exp_lgdplus}}
\tabcolsep=7pt
\setlength{\aboverulesep}{0pt}
\setlength{\belowrulesep}{0pt}
\begin{tabular}{l|c|c|c}
\toprule
\makebox[0.119\textwidth][c]{Methods}	&
\makebox[0.08\textwidth][c]{AGD}	&
\makebox[0.08\textwidth][c]{LGD}	&
\makebox[0.08\textwidth][c]{LGD$+$}	\\
\midrule
{\makebox[0.119\textwidth][c]{$Sim_{Avg}$ (CLIP)}}
	&0.8444		&0.8589		&0.8623\\
\midrule
{\makebox[0.119\textwidth][c]{$Sim_{Best@9}$ (CLIP)}}
	&0.8867		&0.9038		&0.9046\\
\midrule
{\makebox[0.119\textwidth][c]{$AUC_{Avg}$ (CLIP)}}
	&0.8394		&0.8539		&0.8574\\
\midrule
{\makebox[0.119\textwidth][c]{$AUC_{Best@9}$ (CLIP)}}
	&0.8819		&0.8987		&0.8999\\
\toprule
\end{tabular}
\end{table}

\subsubsection{Analysis of Weaknesses}\

Admittedly, the present work still discusses a challenging new task, although we have previously discussed the potential.
Figure \ref{fig15_exp_bad} shows some bad cases.
For example, in the first case on the left, the event that ``the cat jumps up in pain'' is successfully presented.
However, it depends on another event, for instance, ``the train needs to leave'', which must happen immediately as time passes but is not generated.
In particular, it seems that extracting all the changes from the vast amount of image information is still a difficult problem to solve.

Meanwhile, the second case on the left shows that the model still learns biases.
It looks like the model will think of ``light'' when it sees ``lamp'' instead of the apparent cue of ``not properly secured''.

The first example on the right shows the model's confusion about attributes or entities.
We want to see how ``greyhound'' expresses love, not the ``brown dog''.

The last example on the right shows that multimodal reasoning capabilities are still relevant for further research.
Obviously, the model understands that ``Tom'' wants to ``use a weapon'' to attack ``Jerry'', but fails to find the ``knife'' on the table successfully and generates a weird weapon that does not exist.

\subsection{Limitations and Future Work}

In this study, we propose a new image generation task, called VQAI, which aims to investigate the reasoning and generation of causal content for images. 
However, such a task is inherently accompanied by the problem of content diversity, which poses a challenge for the evaluation of the methods.
In this study, we crudely evaluated the proposed method based on CLIP score versus manual scoring. 
However, we argue that it is necessary to develop a more rational automated evaluation matrix as soon as possible.

\newpage
\bibliography{sample-base}

\end{document}